\documentclass[10pt,twocolumn,letterpaper]{article}

\usepackage{wacv}
\usepackage{times}
\usepackage{epsfig}
\usepackage{graphicx}
\usepackage{amsmath}
\usepackage{amssymb}
\usepackage{color}
\usepackage{xspace}
\usepackage{subfig}
\usepackage{float}
\usepackage{contour}
\usepackage{xkeyval}
\usepackage{diagbox}
\usepackage{comment}

\usepackage[pagebackref=false,breaklinks=true,letterpaper=true,colorlinks,urlcolor=blue,citecolor=blue,linkcolor=blue,bookmarks=false]{hyperref}

\usepackage{lipsum}

\wacvfinalcopy 

\def\wacvPaperID{0323} 

\ifwacvfinal\fi
\begin{document}

\definecolor{darkgreen}{RGB}{0,127,0}
\definecolor{darkblue}{RGB}{0,0,127}
\definecolor{darkmagenta}{RGB}{127,0,127}
\definecolor{darkred}{RGB}{127,0,0}
\definecolor{darkcyan}{RGB}{0,127,127}
\definecolor{darkyellow}{RGB}{255,207,0}

\newcommand{\mv}[1]{\mathbf{#1}}
\newcommand{\bd}[1]{\textbf{#1}}
\newcommand{\ud}[1]{#1}
\def\eg{\emph{e.g.}} \def\Eg{\emph{E.g.}}
\def\ie{\emph{i.e.}} \def\Ie{\emph{I.e.}}
\def\cf{\emph{c.f.}} \def\Cf{\emph{C.f.}}
\def\etc{\emph{etc.}} \def\vs{\emph{vs.}}
\def\wrt{w.r.t.} \def\dof{d.o.f.}
\def\etal{\emph{et al.}\xspace}
\def\supmat{\emph{sup.mat.}\xspace}
\def\Fig{Fig.\xspace}

\newcommand\blfootnote[1]{%
  \begingroup
  \renewcommand\thefootnote{}\footnote{#1}%
  \addtocounter{footnote}{-1}%
  \endgroup
}

\newcommand{\wh}{\color{white}} 
\contourlength{.5pt}
\newcommand{\figlabel}[1]{\sffamily\bfseries\wh\scriptsize\contour{black}{#1}} 
\makeatletter
\newlength{\sfp@hseplen}\newlength{\sfp@vseplen}
\define@cmdkey{subfigpos}[sfp@]{vsep}[0.6\baselineskip]{\setlength{\sfp@vseplen}{\sfp@vsep}}
\define@cmdkey{subfigpos}[sfp@]{hsep}[2.5pt]{\setlength{\sfp@hseplen}{\sfp@hsep}}
\newcommand{\subfigimg}[3][,]{%
  \setkeys{Gin,subfigpos}{vsep,hsep,#1}
  \setbox1=\hbox{\includegraphics{#3}}
  \leavevmode\rlap{\usebox1}
  \rlap{\hspace*{8pt}\raisebox{\dimexpr\ht1-7pt}{\figlabel{#2}}}
  \phantom{\usebox1}
}
\makeatother

\newif\ifdrafting
\draftingtrue 

\ifdrafting
  \newcommand{\OG} [1] {\textcolor{darkblue}{[OG: #1]}} 
  \newcommand{\RZ} [1] {\textcolor{darkgreen}{[RZ: #1]}} 
  \newcommand{\DS} [1] {\textcolor{darkmagenta}{[DS: #1]}} 
  \newcommand{\MY} [1] {\textcolor{darkred}{[MY: #1]}} 
  \newcommand{\checkthis}[1]{\textcolor{darkcyan}{[check this $\rightarrow$] #1}} 
\else
  \newcommand{\OG} [1] {}
  \newcommand{\RZ} [1] {}
  \newcommand{\DS} [1] {}
  \newcommand{\MY} [1] {}
  \newcommand{\checkthis}[1]{#1}
\fi

\graphicspath{{figures/}}

\title{A Fusion Approach for Multi-Frame Optical Flow Estimation}


\author{\small Zhile Ren \\
\small Georgia Tech
\and
\small Orazio Gallo \hspace{0.1em} Deqing Sun \\
\small NVIDIA
\and
\small Ming-Hsuan Yang \\
\small UC Merced
\and
\small Erik B. Sudderth \\
\small UC Irvine
\and
\small Jan Kautz \\
\small NVIDIA
}

\maketitle
\ifwacvfinal\thispagestyle{empty}\fi

\begin{abstract}
 To date, top-performing optical flow estimation methods only take pairs of consecutive frames into account.
While elegant and appealing, the idea of using more than two frames has not yet produced state-of-the-art results.
We present a simple, yet effective fusion approach for multi-frame optical flow that benefits from longer-term temporal cues.
Our method first warps the optical flow from previous frames to the current, thereby yielding multiple plausible estimates.
It then fuses the complementary information carried by these estimates into a new optical flow field.
At the time of writing, our method ranks first among published results in the MPI Sintel and KITTI 2015 benchmarks.
%
Our models will be available on \url{https://github.com/NVlabs/PWC-Net}. 
\blfootnote{Part of the work was done while Zhile Ren was an intern at NVIDIA.}
\end{abstract}


\section{Introduction}\label{sec:intro}
\begin{figure}[t]
\centering
\begin{tabular}{cc}
\includegraphics[width=0.24\textwidth]{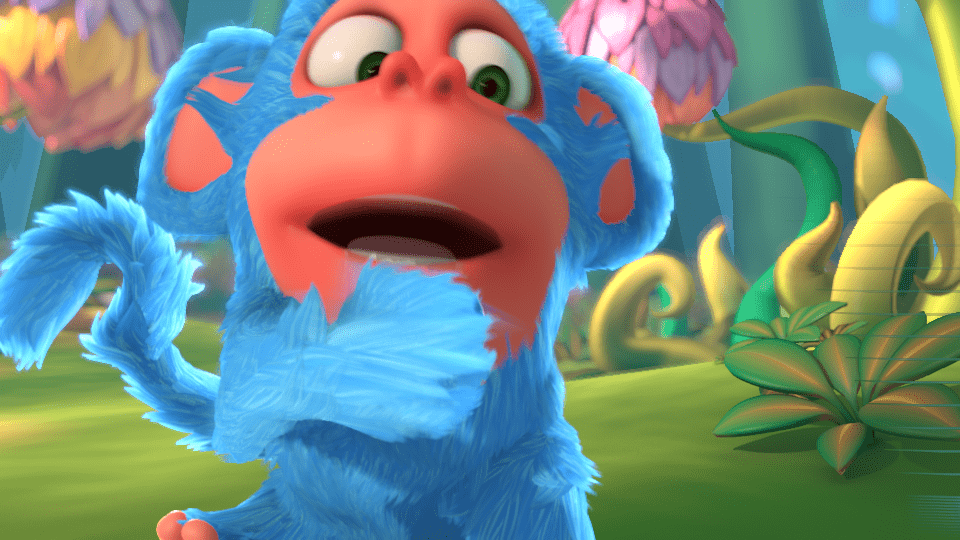} &
\includegraphics[width=0.24\textwidth]{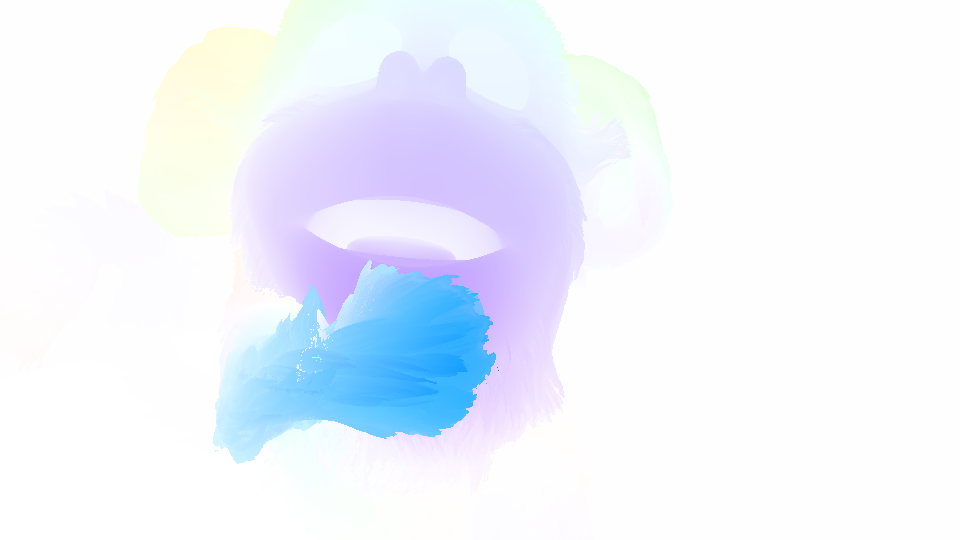} \\
Input & Ground Truth \\
\includegraphics[width=0.24\textwidth]{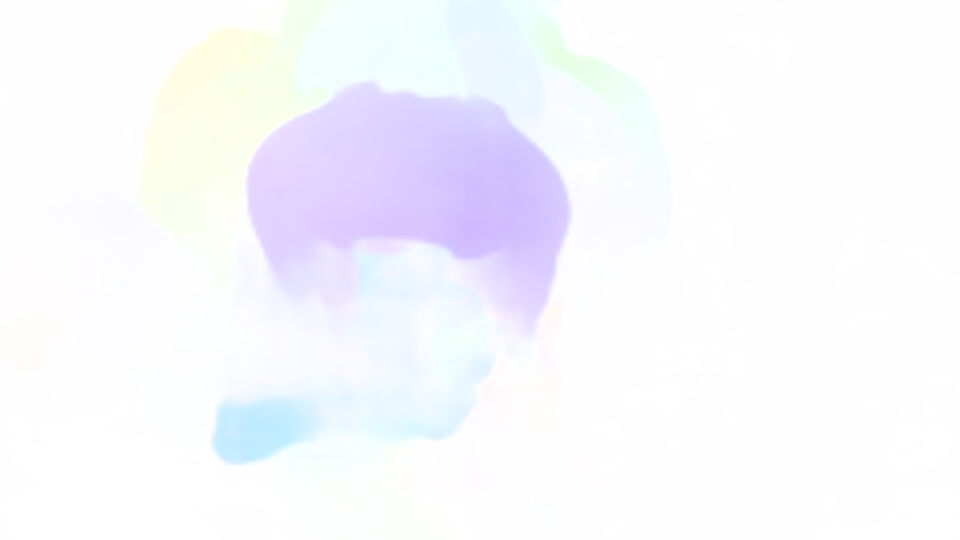} &
\includegraphics[width=0.24\textwidth]{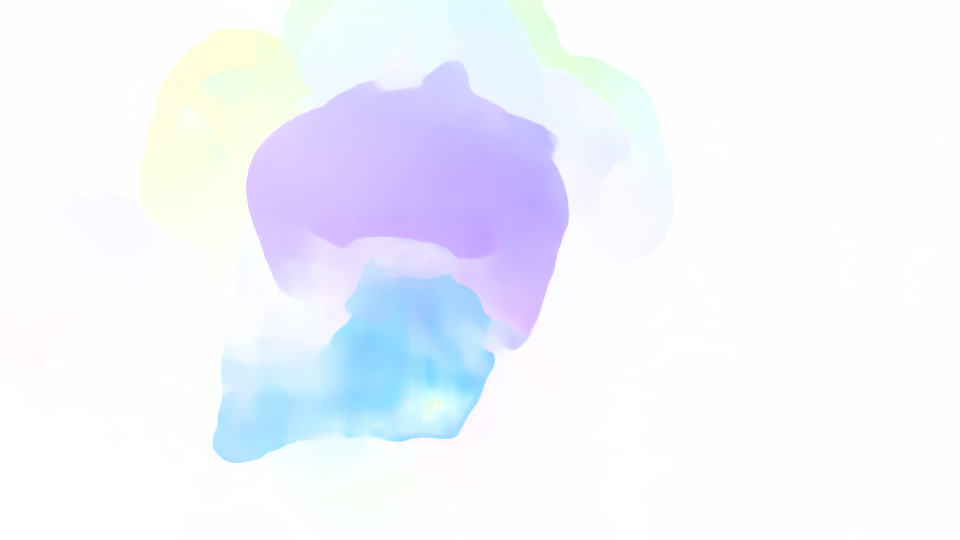} \\
PWC-Net~\cite{sun2017pwc} & PWC-Fusion 
\end{tabular}
\caption{Given a scene with challenging motions, even state-of-the-art two-frame optical flow methods fail to predict the correct motion. 
Our flow fusion approach offers an effective method to incorporate temporal information and improve the predicted flow.
}

\label{fig:teaser}
\end{figure}

Optical flow estimation is an important low-level vision task with numerous applications. 
Despite recent advances, it is still challenging to account for 
complicated motion patterns, as shown in Figure~\ref{fig:teaser}.
At video rates, even such complicated motion patterns are smooth for longer than just two consecutive frames.
This suggests that information from frames that are adjacent in time could be used to improve optical flow estimates.

Indeed, numerous methods have been developed that either impose temporal smoothness of the flow~\cite{Black:1991:RDME,Brox:2004:HAO}, or explicitly reason about the trajectories of pixels across multiple frames~\cite{Sand:2008:PV}.
Despite the fact that multiple frames carry additional information about scene motion, none of the top three 
optical flow algorithms on the major benchmark datasets uses more than two frames~\cite{Butler:eccv:2012,Menze2015CVPR}.

This may be due to the fact that motion and its statistics do change over time. 
If the optical flow field changes dramatically, the visual information contained in longer frame sequences may be less useful, and, in fact, potentially detrimental~\cite{volz2011modeling}.
The ability to decide when visual information from past frames is useful 
is paramount to the success of multi-frame optical flow algorithms.
Some early methods account for the temporal dynamics of motion using a Kalman filter~\cite{elad1998recursive,chin1994probabilistic}. 
More recent approaches attenuate the temporal smoothness strength when a sudden change is detected~\cite{salgado2007temporal,volz2011modeling}.

We observe that, for some types of motion and in certain regions, past frames may carry more valuable information than recent ones, even if the optical flow changes abruptly---as is the case of occlusion regions and out-of-boundary pixels. 
Kennedy and Taylor~\cite{kennedy2015optical} also leverage this observation, and select which \emph{one} of multiple flow estimates from adjacent frames is the best for a given pixel. 
Rather than simply attenuating the contribution of past frames or making an error-prone binary selection, 
we propose a method to \emph{fuse} the available information.
We warp multiple optical flow estimations from the past to the current frame. Such estimates represent plausible candidates of optical flow for the current frame, which we can fuse with a second neural network module.
This approach offers several advantages. 
First, it allows to fully capitalize on motion information from past frames, in particular when this offers a better estimate than the current frame.
Furthermore, to optimally fuse the different optical flow estimates, our method can use information from a large neighborhood.
Since our fusion network is agnostic to the algorithm that generates the pair-wise optical flow estimations, any standard method can be used as an input, making our framework flexible and straightforward to upgrade when improved two-frame algorithms become available.
Finally, if the underlying optical flow algorithm is differentiable, as is the case for the current state-of-the-art methods, our approach can be trained end-to-end.
We show that the proposed algorithm outperforms published state-of-the-art,  two-frame optical flow methods by significant margins on the KITTI~\cite{Geiger:2012:KITTI} and  Sintel~\cite{Butler:eccv:2012} benchmark datasets. 
%
To further validate our results, we present alternative baseline approaches incorporating recurrent neural networks with the state-of-the-art deep-learning optical flow estimation methods, 
and show that significant performance gain can be achieved by using the fusion approach. 

\section{Related Work}\label{sec:related}

Since the seminal work of Horn and Schunck~\cite{Horn:1981:DO}, a rich literature of variational approaches has flourished (\eg,~\cite{Sun:IJCV:2014,Brox:LDOF:2011,EpicFlow}). 
Given the large number of publications in this area, we refer the reader to~\cite{Sun:IJCV:2014,Baker:2011:DEO} for a more complete survey focusing on two-frame optical flow methods.
Since the optical flow field between two frames captures the displacement of pixels over time, exploiting longer time sequences may help in ambiguous situations such as occluded regions~\cite{Black:2000:DTMB,Feldman:2008:OD,Sun:2012:LSOT}. 
Despite the long history of multi-frame optical flow estimation~\cite{Black:1990:MDMT,Murray:1987:SSVM}, few methods have concretely demonstrated the benefits of multi-frame reasoning.
In the following, we briefly review these multi-frame optical flow methods.

\paragraph{Temporal smoothing.}
The most straightforward way to leverage information across more than a pair of frames is to encourage temporal smoothness of the optical flow.
Nagel imposes a spatio-temporal oriented smoothness constraint that is enforced along image edges, but relaxed across them~\cite{nagel1990extending}.
Weickert and Schn{\"o}rr also propose to combine temporal and spatial smoothness into a variational penalty term, but relax this constraint at flow discontinuities~\cite{weickert2001variational}. 
Zimmer~\etal propose a principled way to attenuate the temporal discontinuity: they weigh the spatio-temporal smoothness based on how well it predicts subsequent frames~\cite{zimmer2011optic}.
Instead of using variational temporal smoothness terms, optical flow estimation can also be cast as a Kalman filtering problem~\cite{elad1998recursive,chin1994probabilistic}.

\paragraph{Explicit trajectory reasoning.}
Enforcing temporal smoothness of the optical flow is helpful, but it neglects higher level information such as the trajectory of objects, which may cause large and small displacements to coexist.
A different, arguably more principled solution is to warp the optical flow estimation from previous frames to the current. 
Assuming the acceleration of a patch to be constant, Black and Anandan warp the velocity predicted from the previous frame to the current one, obtaining $\hat{\mv{w}}(t)$~\cite{black1991robust}. 
They formulate the temporal constraint as $f(\hat{\mv{w}}(t) - \mv{w}(t))$, where $\mv{w}(t)$ is the velocity estimated at the current frame.
One shortcoming of this approach is that information is only propagated forward.
Werlberger~\etal tackle this problem by using the forward flow from $t-1$ and the backward flow from $t+1$ in the brightness constancy term for frame $t$~\cite{werlberger2009anisotropic}.
On the other hand, Chaudhury and Mehrotra pose optical flow estimation as the problem of finding the shortest path with the smallest curvature between pixels in frames of a sequence~\cite{chaudhury1995trajectory}.
Similarly, Volz~\etal enforce temporal consistency from past and future frames in two ways: accumulating errors across all frames leads to spatial consistency, and trajectories are also encouraged to be smooth~\cite{volz2011modeling}.

\paragraph{Multi-frame approaches.} Irani observes that the set of plausible optical flow fields across multiple frames lie on a lower dimensional linear subspace and does not require explicitly enforcing temporal or spatial smoothness~\cite{irani1999multi}. However, the observation applies to rigid scenes and cannot be directly applied to dynamic scenes.
Noting that the optical flow may not be temporally consistent, Sun~\etal propose to impose consistency of the scene structure, \ie, layered segmentation~\cite{Wang:1994:RMIL}, over time, and combine motion estimation and segmentation in a principled way~\cite{Sun:2012:LSOT}. 
Wulff~\etal decompose a scene into rigid and non-rigid parts, represent the flow in the rigid regions via depth, and regularize the depth over three frames~\cite{Wulff2017Optical}. Both these formulations, unfortunately, lead to a large optimization problem that is computationally expensive to solve.

\paragraph{Flow fusion.}
Fusion is a widely used approach to integrate different estimates of the flow fields into a more accurate one. Lempitsky~\etal generate flow candidates by applying Lucas and Kanade~\cite{Lucas:1981:LK} and Horn and Schunck~\cite{Horn:1981:DO} with different parameters settings to the same two input frames, and fuse these candidates via energy minimization~\cite{Lempitsky:2010:FM}. 
To deal with small and fast moving objects, Xu~\etal obtain flow candidates by feature matching and fuse them with the estimates by a variational approach in a coarse-to-fine pyramid~\cite{Xu:2012:MDP}. 
Mac Aodha~\etal generate the flow candidates by applying a variety of different algorithms and fuse them using a random forest classifier~\cite{MacAodha:2010:SVC}.  These methods are all designed for two input frames and do not consider multi-frame temporal cues. Kennedy and Taylor acquire several flow estimates from adjacent frames and fuse them using a random forest classifier~\cite{kennedy2015optical}. However, their system requires an additional post-processing step and is not end-to-end trainable. {One limitation of the discrete fusion approach is that a pixel only picks up  flow vectors from the flow candidates at the same spatial location, although neighboring regions may provide better candidates.}
Recently, Maurer and Bruhn proposed a fusion-based flow estimation method that is closely related to our approach, but the algorithm does not outperform two-frame methods on all datasets~\cite{maurer2018proflow}.
\vspace{-1.5em}
\paragraph{Deep learning-based two-frame methods.}
The recent success of deep learning inspired several approaches based on neural networks, which can either replace parts of the traditional pipeline~\cite{Xu2017Accurate,Bailer2017CNN}, or can be used in an end-to-end fashion~\cite{Dosovitskiy:2015Flownet,Ilg:2016:Flownet2,Ranjan:2016:SpyNet,sun2017pwc}. 
The top-performing deep-learning optical flow algorithms are still based on two frames and do not consider additional temporal information. In this paper, we apply our fusion approach to both FlowNetS and PWC-Net and show that temporal information leads to significant improvements to both methods.

\section{Proposed Model}\label{sec:model}

\begin{figure*}[!ht]
	\centering
	\includegraphics[width=0.85\linewidth,trim={50pt 200pt 65 165pt}, clip]{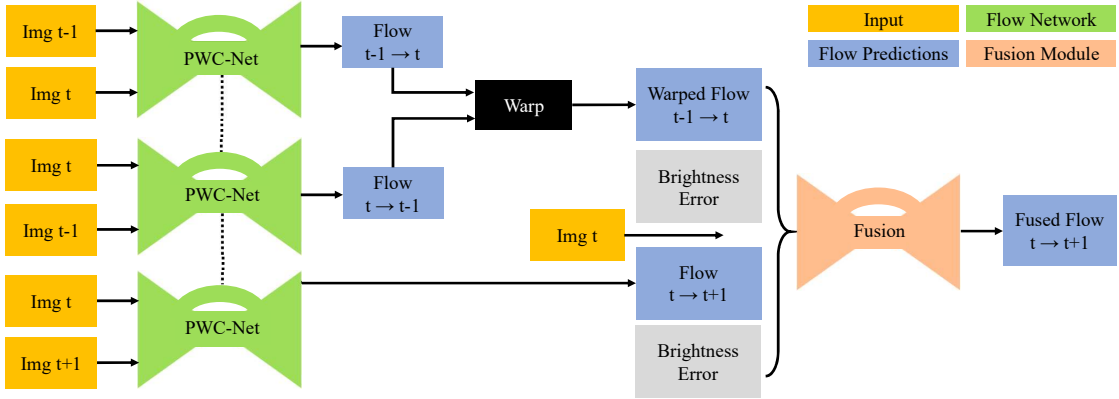}
	\caption{Architecture of the proposed fusion approach for three-frame optical flow estimation. The dashed line indicates that the PWC-Nets share the same weights. PWC-Net can be replaced with other two-frame flow methods like FlowNetS.}\label{fig:system}
\end{figure*}
\begin{figure*}[!ht]
\begin{tabular}{cccc}
\subfigimg[width=0.24\linewidth]{GT Flow t to t+1}{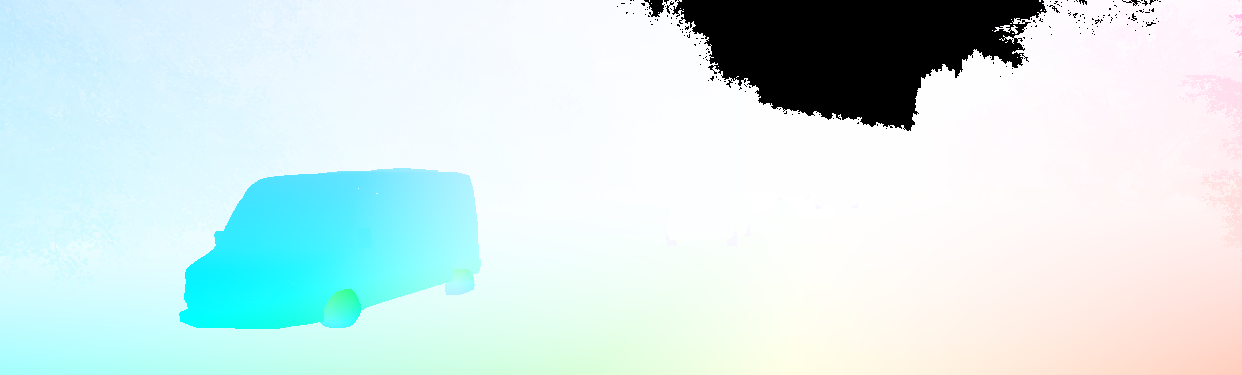} &
\subfigimg[width=0.24\linewidth]{Flow t to t+1}{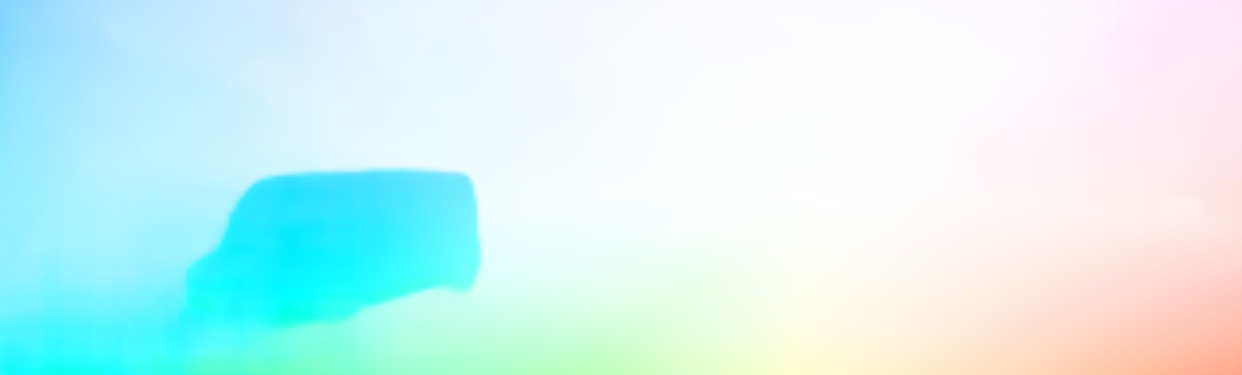} &
\subfigimg[width=0.24\linewidth]{Warped Flow t to t+1}{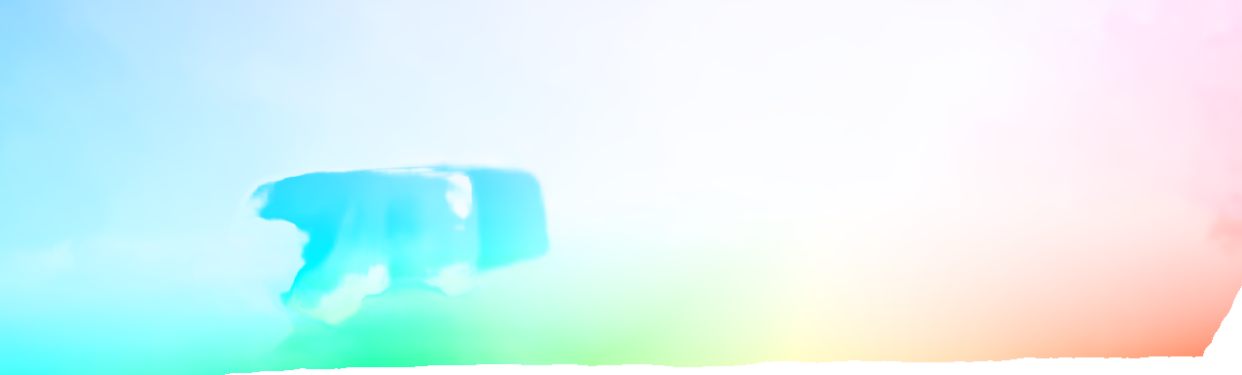} &
\subfigimg[width=0.24\linewidth]{Fused Flow}{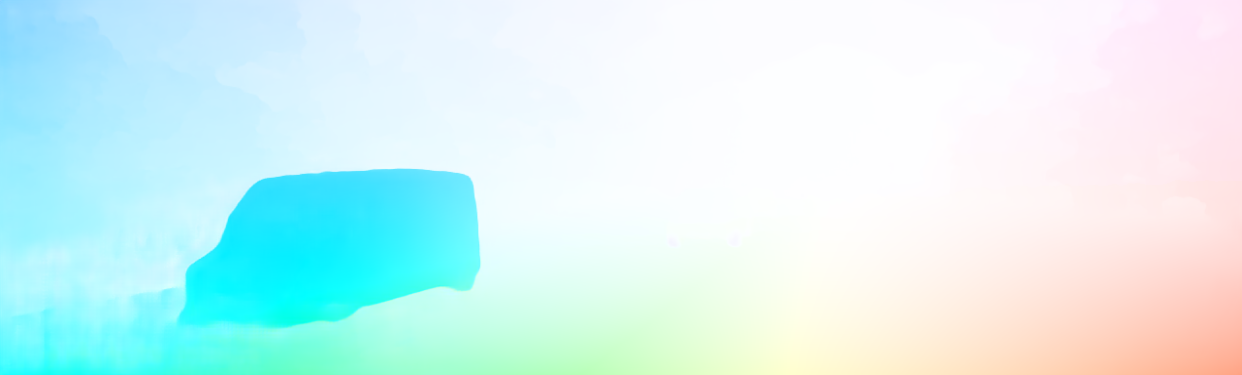} \\
\subfigimg[width=0.24\linewidth]{Img t}{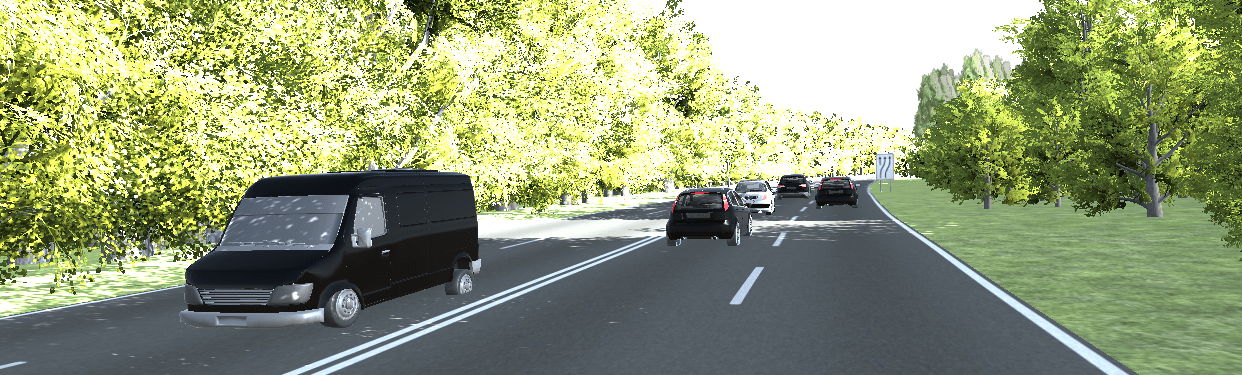} &
\subfigimg[width=0.24\linewidth]{Brightness Error}{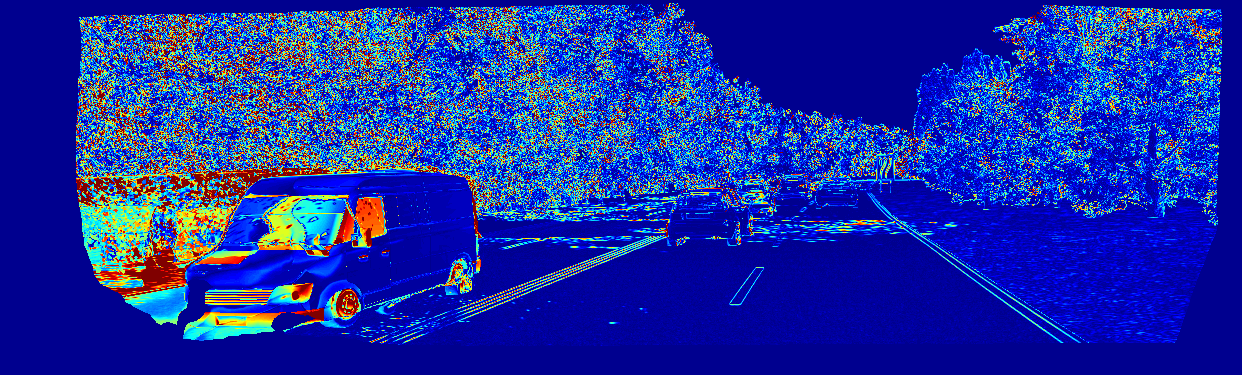} &
\subfigimg[width=0.24\linewidth]{Brightness Error}{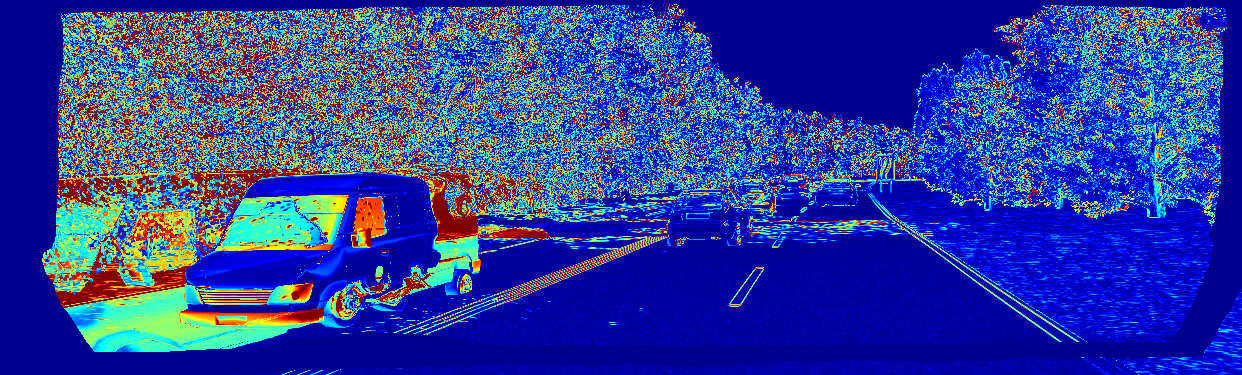} &
\subfigimg[width=0.24\linewidth]{Fusion Indicator Map}{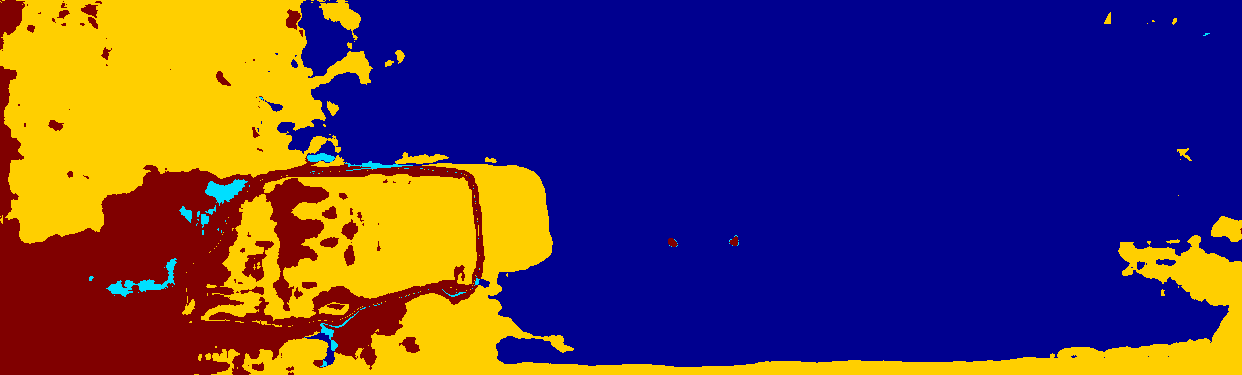} \\
\end{tabular}
\caption{Visualizing the input and behavior of the fusion network. In the fusion indicator map,
we compare the fused flow with two flow candidates. {\color{blue}{Blue}}: both candidates
are similar ($<5$px), {\color{darkyellow}{yellow}}: the fused flow is similar to $\mv{w}_{t \rightarrow t+1}$, and 
{\color{cyan}{cyan}}: the fused flow is similar to $\widehat{\mv{w}}_{t \rightarrow t+1}$. 
{\color{red}{Red}}: the fused flow is different from all flow candidates (mostly occluded regions).}
\label{fig:vis_fusion}
\vspace{-1em}
\end{figure*}

For presentation clarity, we focus on three-frame optical flow estimation. Given three input frames $\mv{I}_{t\!-\!1}$, $\mv{I}_{t}$, and $\mv{I}_{t\!+\!1}$, our aim is to estimate the optical flow from frame $t$ to frame $t+1$, $\mv{w}^f_{t \rightarrow t+1}$. The superscript `$f$' indicates that it fuses information from all of the frames, as opposed to $\mv{w}_{t \rightarrow t+1}$, which only uses $\mv{I}_{t}$ and $\mv{I}_{t\!+\!1}$. 

We hypothesize that the ability to leverage the information from $\mv{w}_{t-1 \rightarrow t}$ can help in several ways.
First, it can act as a regularizer, since the optical flow between consecutive pairs of frames changes smoothly in most regions.
Perhaps more importantly, however, $\mv{w}_{t-1 \rightarrow t}$ can offer information that is complementary to that of $\mv{w}_{t \rightarrow t+1}$, especially for pixels that are occluded or out of boundary in the current frame or the next.
Finally, given those two observations, the optimal estimate for $\mv{w}^f_{t \rightarrow t+1}$ may also benefit from information from local neighborhoods in both images.

In this section, we begin by verifying that temporal information indeed helps and by determining the image regions where it is most informative.  We then introduce the proposed multi-frame fusion architecture, and discuss two deep learning baselines for multi-frame optical flow estimation.

\subsection{Temporal Information: An ``Oracle'' Study}\label{sec:verifying}

We evaluate the benefits of temporal information using the virtual KITTI~\cite{Gaidon:Virtual:CVPR2016} and Monkaa~\cite{Mayer:2016:Large} datasets. We split each video sequence into three-frame mini-sequences and use two-frame methods to estimate three motion fields,  $\mv{w}_{t \rightarrow t+1}$,  $\mv{w}_{t-1 \rightarrow t}$, and $\mv{w}_{t \rightarrow t-1}$. We first warp these optical flow fields into a common reference frame. Specifically, we backward warp $\mv{w}_{t-1\rightarrow t} $ with $\mv{w}_{t \rightarrow t-1}$:
\begin{equation}\label{eq:warp}
\widehat{\mv{w}}_{t \rightarrow t+1} = \mathcal{W} (\mv{w}_{t-1 \rightarrow t}; \mv{w}_{t \rightarrow t-1}),
\end{equation}
where $\mathcal{W} (\mv{x}; {\mv{w}})$ denotes the result of warping the input $\mv{x}$ using the flow field $\mv{w}$.

We then compare warped and current flows with the ground truth at every pixel, and select the one closer to the ground truth to obtain an ``oracle'' flow field. We test the ``oracle'' flow fields for two flow methods, FlowNetS and PWC-Net. As shown in Table~\ref{table:epe_vkitti_monkaa},  the ``oracle'' flow fields are more accurate than the current flow estimated by two-frame methods, particularly in occluded and out-of-boundary regions. The results confirm that the warped previous optical flow provides complementary information and also suggest a straightforward mechanism for multi-frame flow estimation.

\subsection{Temporal FlowFusion}
In a nutshell, our architecture consists of two neural networks. The first computes the optical flow between adjacent frames. These optical flow estimates are then warped onto the same frame, producing two optical flow candidates. A second network then fuses these two. Both networks are pre-trained and then fine-tuned in an end-to-end fashion, see Figure~\ref{fig:system}.

More specifically, our system first uses a two-frame method, such as PWC-Net~\cite{sun2017pwc}, to estimate three flow fields $\mv{w}_{t \rightarrow t+1}$, $\mv{w}_{t-1 \rightarrow t}$, and $\mv{w}_{t \rightarrow t-1}$.
Any other two-frame method could be used in place of PWC-Net. However, we pick PWC-Net as it currently is the top-performing optical flow method on both the KITTI 2015 and Sintel final benchmarks. 

Now we have two candidates for the same frame: $\widehat{\mv{w}}_{t \rightarrow t+1}$ and $\mv{w}_{t \rightarrow t+1}$.
Thanks to the warping operation, these two optical flow should be similar (identical in regions where the velocity is constant, the three optical flow estimations are correct, and there are no occlusions).
To fuse these, we take inspiration from the work of Ilg~\etal who perform optical flow fusion in the spatial domain for two-frame flow estimation~\cite{Ilg:2016:Flownet2}.
Specifically, they propose to estimate two optical flows, one capturing large displacements and one capturing small displacements, and to fuse them using a U-Net architecture. The input to their fusion network are the flow candidates, their brightness constancy error, flow magnitude, and the first frame. The  magnitude of the flow is an important feature for their network to distinguish large and small motion. 


We extend this approach to the temporal domain. Our fusion network takes two flow estimates $\widehat{\mv{w}}_{t \rightarrow t+1}$ and $\mv{w}_{t \rightarrow t+1}$, the corresponding brightness constancy errors
\begin{eqnarray}
E_{\widehat{\mv{w}}} &=& |\mv{I}_t - \mathcal{W} ({\mv{I}}_{t+1}; \widehat{\mv{w}}_{t \rightarrow t+1})|,~~\text{and}\\
E_\mv{w} &=& |\mv{I}_t - \mathcal{W} ({\mv{I}}_{t+1}; \mv{w}_{t \rightarrow t+1})|,
\end{eqnarray}
as well as the current frame $\mv{I}_t$. 
Since we are not trying to fuse optical flow fields that were estimated at different scales, the magnitude of the flow, which Ilg~\etal employ, may not be useful and potentially detrimental.
Other inputs could be used, such as a measure of smoothness between the previous flow, $\mv{w}_{t-1 \rightarrow t}$, and the current, $\mv{w}_{t \rightarrow t+1}$.
However, we found that the simpler and lighter-weight input we use works best in our experiments. We provide more analysis of the behavior of our fusion network in Section~\ref{sec:experiments}.

Note that, while we take inspiration from FlowNet2~\cite{Ilg:2016:Flownet2} for the fusion step, our approach and our goal differ from theirs conceptually: we explore and leverage the benefits of temporal information.

\subsection{Deep-Learning Baseline Methods}\label{sec:grus}

To the best of our knowledge, our flow fusion method is the first realtime deep learning-based method 
to fuse temporal information in the context of optical flow. 
To evaluate the performance of our approach, we build on standard deep-learning optical flow algorithms, and propose two baseline methods.
We extensively validate our method in Section~\ref{sec:experiments} and find that our fusion network, despite its simplicity, works consistently better than those baselines.
Here, we first briefly review the methods upon which we build, and then introduce the proposed baselines.

\begin{figure}[!ht]
	\centering
	\includegraphics[width=0.5\linewidth]{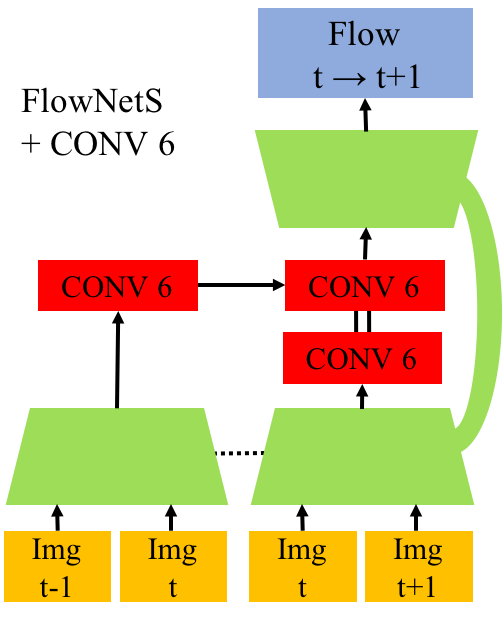}
	\\
	 \includegraphics[width=0.8\linewidth]{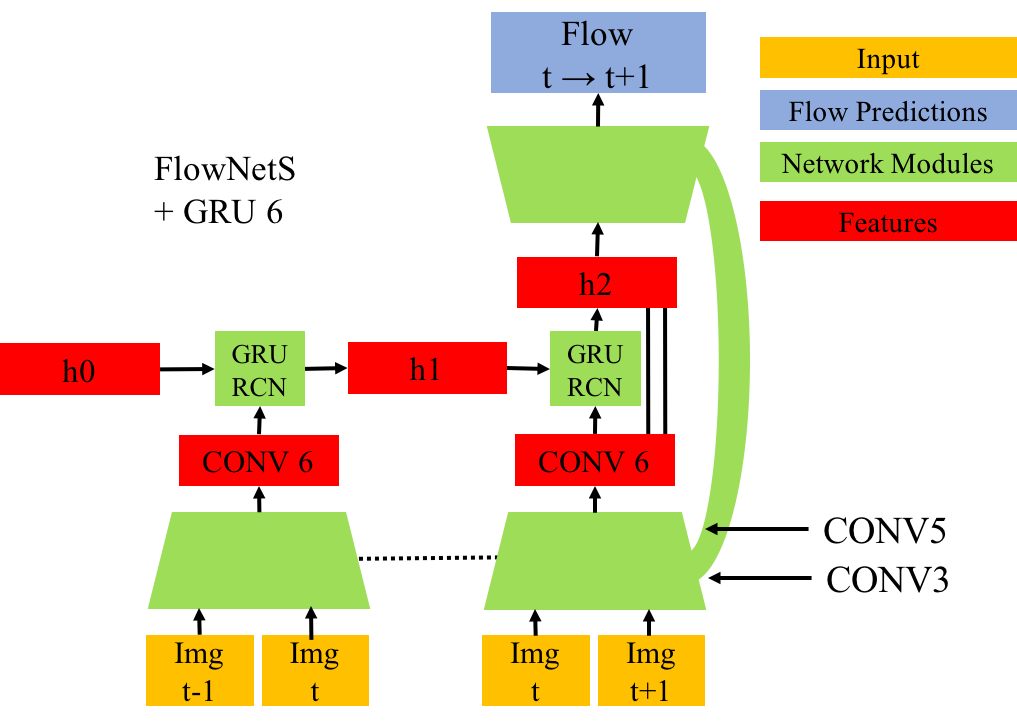}
	\caption{Baseline network structures for modeling temporal information. FlowNetS++ (top)
	         simply concatenates the encoded features across different frames, and 
	         FlowNetS $+$ GRU6 (bottom) propagates the encoded features through GRU-RCN units. The dotted lines indicate that two sub-networks share the same weights, while the double vertical lines denote the feature concatenation operation.}
	\label{fig:network_time}
\vspace{-1em}	
\end{figure}

\noindent{\bf FlowNetS++}~~The most widely used deep network structure for optical flow prediction is
FlowNetS~\cite{Dosovitskiy:2015Flownet}, which is a standard U-Net structure. The most natural way to extend
this model to multi-frame prediction tasks is to copy the encoded features from the previous pair of images to
the current frame (Figure~\ref{fig:network_time}, left). The inner-most high-dimensional feature of the 
previous frame may carry instructive information to guide motion prediction at the next frame. We call this
FlowNetS++.

\noindent{\bf FlowNetS $+$ GRU}~~In deep learning, temporal information
is often captured with Recurrent Neural Networks (RNNs). In recent years, the popular LSTM~\cite{hochreiter1997long} 
and GRU~\cite{chung2014empirical} network architectures have been applied to a variety of vision tasks, such as action recognition~\cite{Martinez:CVPR:2017}
and segmentation~\cite{byeon2015scene}. GRUs have a comparatively smaller model size and the convolutional extension 
GRU-RCN~\cite{ballas2015delving} is widely used to extract abstract representations from videos. We propose to use
GRU-RCN to propagate encoded features in previous frames through time in a GRU-RCN unit 
and introduce a network structure, which we dub FlowNetS $+$ GRU (Figure~\ref{fig:network_time}, right). We preserve the overall
U-Net structure and apply GRU-RCN units at different levels of the encoder with different spatial resolutions. 
Encoded features at the sixth level are the smallest in resolution. 

\noindent{\bf PWC-Net $+$ GRU}~~We can apply a similar strategy to PWC-Net~\cite{sun2017pwc}, a recently introduced network that achieves excellent performance 
for two-frame optical flow prediction tasks. The network first feeds two images into separate siamese networks, 
which consist of a series of convolutional structures. Then it decodes features and learns
abstract representations at different levels. Similarly to FlowNetS $+$ GRU, we can also feed encoded features at 
different levels to GRU-RCN units and we call it PWC-Net $+$ GRU. A depiction of this second network structure is omitted for clarity.

\section{Experimental Results}\label{sec:experiments}


We use two different architectures as building blocks:
FlowNetS~\cite{Dosovitskiy:2015Flownet} for its widespread adoption, 
and PWC-Net~\cite{sun2017pwc} for its efficiency and performance on standard benchmarks.
We pretrain each of these on the ``FlyingChairs'' dataset~\cite{Dosovitskiy:2015Flownet} first, and then use the learned
weights to initialize the different network structures for multiple frame optical flow estimations.
For consistency among different multi-frame algorithms, 
we use three frames as inputs, and report ablation studies on 
the virtual KITTI dataset~\cite{Gaidon2016Virtual} and the Monkaa dataset~\cite{Mayer:2016:Large}. Finally, we
report the performance of our algorithms on the two public benchmarks KITTI~\cite{Geiger:2012:KITTI} and Sintel~\cite{Butler:eccv:2012}.

\setlength\tabcolsep{3pt} 
\begin{table*}[!ht]
\scriptsize
\centering
\caption{Ablation study on the \emph{validation} set of the virtual KITTI dataset (top) and Monkaa dataset (bottom). Training a 
temporal network structure helps improve two-frame
optical flow predictions, but the fusion approach has significantly lower end-point-error (EPE). Inside means pixels move within the image boundary and Outside means pixels that move out of the image boundaries.}
\label{table:epe_vkitti_monkaa}
\begin{tabular}{c|c|c|c|c|c|c|c|c||c|c|c|c|c|c|c} 
& FlowNetS &FlowNetS++ & GRU 3 &GRU 4 &GRU 5 &GRU 6 & Fusion & Oracle & PWC-Net &GRU 3 &GRU 4 &GRU 5 &GRU 6 & Fusion & Oracle\\ \hline
EPE All &6.12 &5.90 &5.26 &5.40 &5.15 &5.32 & \textbf{5.00} & 4.35 &2.34 &2.17 &2.13 &2.12 &2.16 &\textbf{2.07} &1.80\\ 
EPE Inside& 4.03 &3.87 &3.61 &3.64 &3.58 &3.59 &\textbf{3.14} &2.62& 1.60&1.44 &1.41 &1.40 &1.42 &\textbf{1.37}&1.20\\
EPE Outside& 28.97 &27.57 &23.26 &24.60 &\textbf{22.28} &24.25 &25.15 &21.83&10.43 &10.01 &9.94 &10.02 &9.86 &\textbf{9.71} &7.90\\
EPE Occlusion& 7.44&7.11&\textbf{5.93}&6.27&5.82&6.18&6.14 & 4.63&2.41&2.29&\textbf{2.24}&\textbf{2.24}&2.26&2.27 & 1.82
\end{tabular}
\begin{tabular}{c|c|c|c|c|c|c|c|c||c|c|c|c|c|c|c} 
& FlowNetS &FlowNetS++ & GRU 3 &GRU 4 &GRU 5 &GRU 6 & Fusion& Oracle& PWC-Net & GRU 3 &GRU 4 &GRU 5 &GRU 6 & Fusion& Oracle \\ \hline
EPE All &2.07 &2.06 &2.56 &2.45 &2.34 &2.27 &\textbf{1.97}&1.89&1.19 &1.26 &1.23 &1.27 &1.27 &\textbf{1.18} &1.03\\ 
EPE Inside& 1.91&1.89 &2.37 &2.27 &2.16 &2.09 &\textbf{1.8}&1.75&1.19 &1.26 &1.23 &1.27 &1.27 &\textbf{1.18} &1.03\\ 
EPE Outside&11.47 &\textbf{11.12} &13.13 &12.76 &12.65 &12.43 &11.43 &9.79&8.00 &8.16 &8.11 &8.55 &8.42 &\textbf{7.94}&6.90\\
EPE Occlusion&8.16&8.02&9.17&9.07&8.84&8.71&\textbf{7.89} &7.00&5.83&5.97&5.90&5.80&5.79&\textbf{5.67}& 4.88
\end{tabular}

\end{table*}

\begin{table*}[!ht]
	\scriptsize	
	\hspace{-3em}
		\caption{Experimental results on the MPI Sintel dataset~\cite{Butler:eccv:2012} (left) and KITTI~\cite{Geiger:2012:KITTI} (right). Our PWC-Fusion
		method ranks first among all published methods.}
	\label{table:epe_sintel}
	\begin{tabular}{c|c|c|c|c|c|c|c|c|c} 
		&EPE&Match&Unmatch&d0-10&d10-60&d60-140&s0-10&s10-40&s40+\\ \hline
		PWC-Fusion &\textbf{4.566}&\textbf{2.216}&23.732&\textbf{4.664}&\textbf{2.017}&\textbf{1.222}&0.893&2.902&26.810\\
		PWC-Net &4.596&2.254&\textbf{23.696}&4.781&2.045&1.234&0.945&2.978&\textbf{26.620}\\
		ProFlow &5.015&2.659&{24.192}&4.985&2.185&1.771&0.964&2.989&29.987\\		
		DCFlow &5.119&{2.283}&28.228 &4.665&2.108&1.440&1.052&3.434&29.351\\
		\hspace{-1.3em}FlowFieldsCNN &5.363&2.303&30.313 &4.718&{2.020}&{1.399}&1.032&3.065&32.422\\
		MR-Flow &5.376&	2.818	&26.235&5.109&2.395&1.755&0.908&3.443&32.221\\
		LiteFlowNet &5.381&	2.419&	29.535&	4.090&	2.097&	1.729&\textbf{0.754}&	\textbf{2.747}&	34.722\\
		S2F-IF &5.417&2.549&28.795&4.745&2.198&1.712&1.157&3.468&31.262
	\end{tabular}
	\hspace{0.1em}
	\begin{tabular}{c|c|c|c|c|c|c}
&Fl-all-Occ&Fl-fg-Occ&Fl-bg-Occ&Fl-all-Ncc&Fl-fg-Ncc&Fl-bg-Ncc\\ \hline
PWC-Fusion&\textbf{7.17}&\textbf{7.25}&\textbf{7.15}&\textbf{4.47}&\textbf{4.25}&\textbf{4.52}\\
PWC-Net &7.90&8.03&7.87&5.07&5.04&5.08\\
LiteFlowNet &9.38&{7.99}&9.66&5.49&5.09&5.58\\
MirrorFlow &10.29&17.07&8.93&7.46&12.95&6.24\\
SDF &11.01&23.01&{8.61}&8.04&18.38&5.75\\
UnFlow &11.11&15.93&10.15&7.46&12.36&6.38\\
MRFlow &12.19&22.51&10.13&8.86&17.91&6.86\\
ProFlow &15.04&20.91&13.86&10.15&17.9&8.44
\end{tabular}

\end{table*}

\subsection{Implementation Details}
Traditional optical datasets~\cite{Baker:2011:DEO,Barron:1994:PO} contain a small number of images and thus are not suitable for 
training modern deep network structures. Dosovitskiy~\etal find that accurate models can be trained using computer graphics rendered images and introduce the ``FlyingChairs'' dataset as well as a corresponding learning rate schedule~\cite{Dosovitskiy:2015Flownet}. A follow-up work then investigates a better learning rate schedule for training using multiple datasets~\cite{Ilg:2016:Flownet2}.  We use the same
strategy $S_{\text{long}}$ proposed by Ilg~\etal~\cite{Ilg:2016:Flownet2} to pretrain our two-frame optical flow algorithms using the ``FlyingChairs''  and ``FlyingThings3D'' datasets. 

Since we aim at predicting optical flow from multi-frame inputs, we train on datasets that
offer videos, rather than just pairs of images. The Monkaa dataset consists of $24$ video sequences with $8591$ frames and the virtual KITTI dataset consists of $5$ different scenes with $21,260$ frames. 
For our ablation study, we split both Monkaa and virtual KITTI datasets into non-overlapping training and testing portions. Specifically, Monkaa has $55104$ frames for training and $10944$ frames for validation and virtual KITTI has three sequences ($14600$ frames) for training and two sequences ($5600$ frames) for validation.

After pretraining our Fusion model using Monkaa and virtual KITTI, we fine-tune them using the training set of the Sintel and KITTI benchmarks, and compare our method with state-of-the-art algorithms on the two public benchmarks. When fine-tuning on the Sintel dataset, we use pretrained weights from the Monkaa dataset because both datasets consist of animated cartoon
characters generated from the open source Blender software. For fine-tuning on the KITTI dataset, we use pretrained weights from the virtual KITTI dataset. 
We adopt the robust training loss from Sun~\etal~\cite{sun2017pwc}:
\begin{equation*}
L(\Theta)=\sum_{l=l_0}^{L}\alpha_l\sum_\mv{x}(|\mv{w}_\Theta^l(\mv{x})-\mv{w}_{GT}^l(\mv{x})| +\epsilon)^q+\gamma\lVert\Theta\rVert^2_2.
 \label{eq:robustloss}
\end{equation*}
We follow the notation defined by Sun~\etal~\cite{sun2017pwc} and we refer readers to their paper for more details.
Briefly, $\alpha_l$ is the weight at pyramid level $l$; for smaller spatial
resolutions, we set a higher weight for $\alpha_l$. 
The robustness parameter $q$ controls the penalty for outliers in flow predictions, $\epsilon > 0$ is a small constant, and $\gamma$ determines the amount of regularization.  
Parameters are set as follows:
\begin{equation*}
\begin{aligned}
 \alpha=(0.005,0.01,0.02,0.08,0.32) \\ \epsilon=0.01, q = 0.4, \gamma=0.0004.
 \end{aligned}
\end{equation*}

For multi-frame networks using GRU-RCN units, we use 256-dimensional hidden units.
For fusion networks, the network structure is similar to FlowNet2~\cite{Ilg:2016:Flownet2} except for the first convolution layer, because
our input to the fusion network has different channels. For the single optical flow prediction output by our fusion network, 
we set $\alpha=0.005$ in the loss function and use learning rate $0.0001$ for fine-tuning.
We implement our models using the PyTorch framework~\cite{paszke2017automatic}, including the pretrained models for FlowNetS and PWC-Net.
Our implementation matches the official Caffe implementation of PWC-Net.

\subsection{Ablation Study}
\subsubsection{Two-frame and Multi-frame Methods}
We perform an ablation study of the two-frame and multi-frame methods using the virtual KITTI and Monkaa datasets, as summarized in Table~\ref{table:epe_vkitti_monkaa}. 

FlowNetS++ shows a small but consistent improvement over FlowNetS, in every different region, including pixels that move within (Inside) and out of (Outside) the image boundaries or get occluded. 
The results show the benefit of using temporal information for optical flow, even in the most na{\"i}ve and straightforward way. 
FlowNetS+GRU has a relatively larger improvement compared to FlowNetS++, suggesting the advantage of using better models to capture the dynamics over time. However, there is no obvious winner among architectures augmented by using GRUs at different pyramidal levels. 

\begin{figure*}[ht!]
\begin{tabular}{ccc}

\subfigimg[width=0.32\linewidth]{Input t}{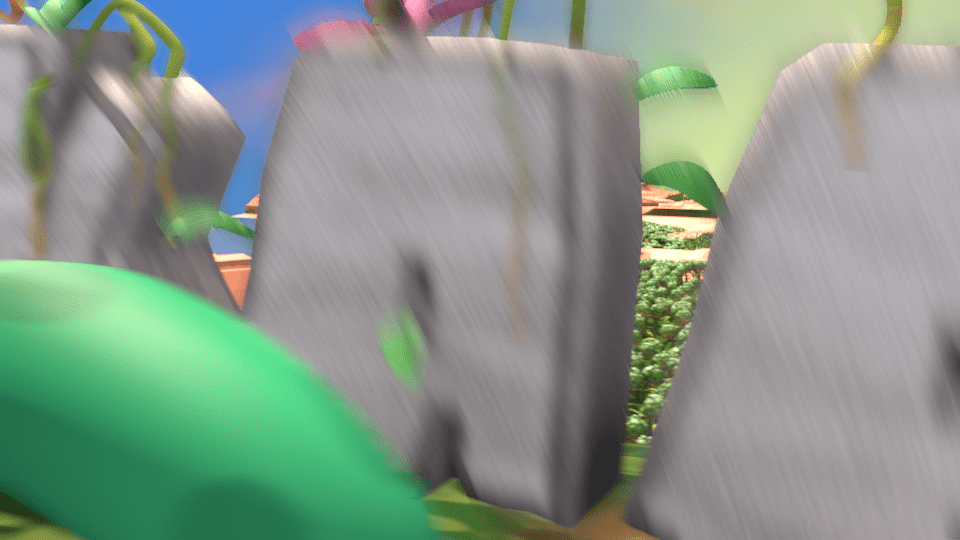} &
\subfigimg[width=0.32\linewidth]{Input t+1}{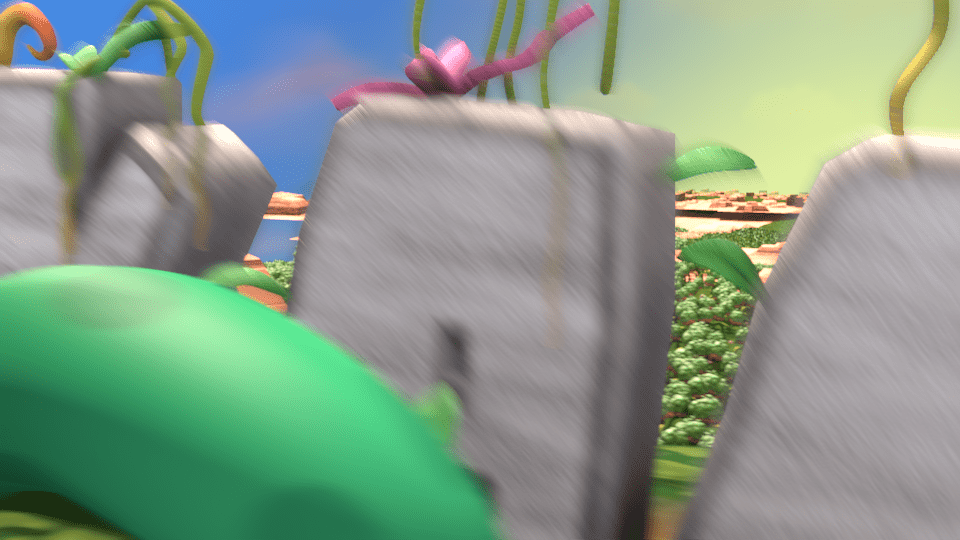} &
\subfigimg[width=0.32\linewidth]{GT Flow}{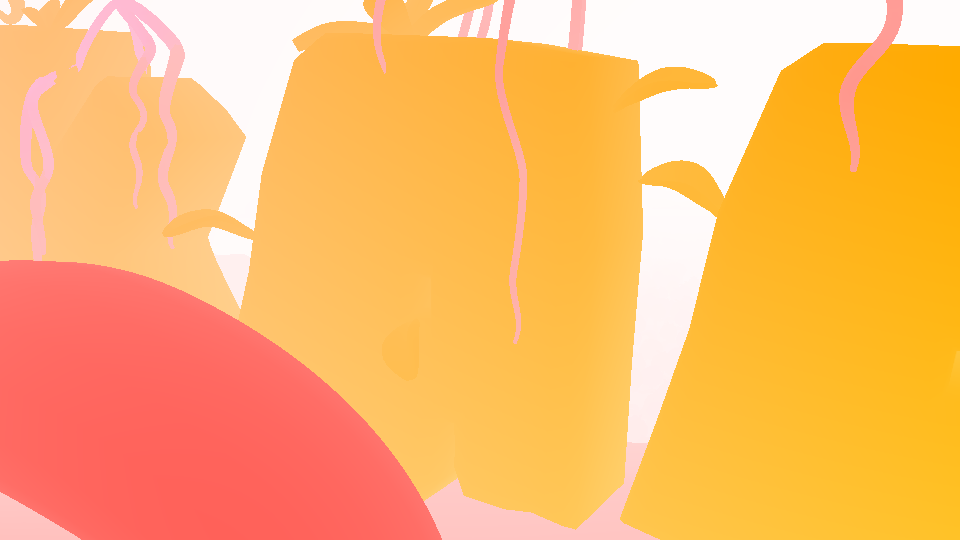} \\
\subfigimg[width=0.32\linewidth]{FlowNetS}{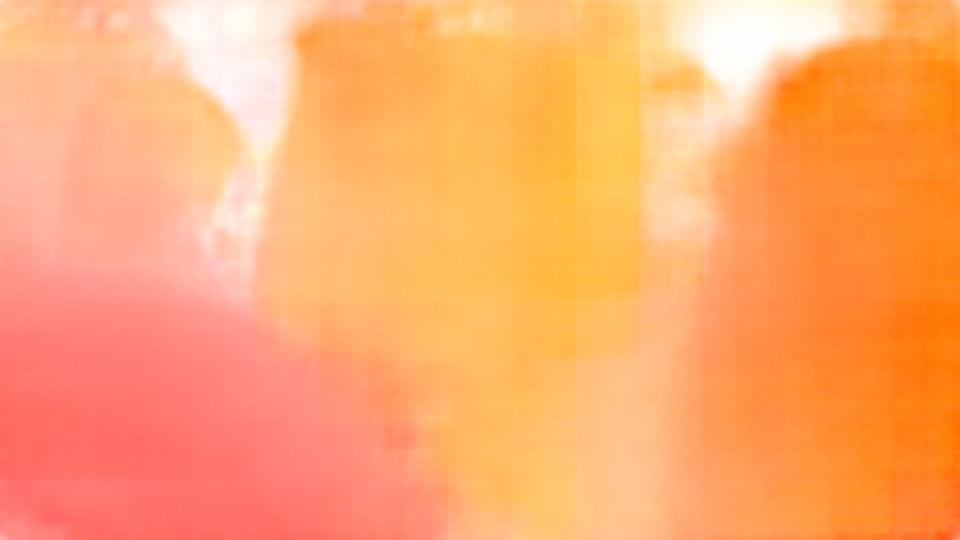} &
\subfigimg[width=0.32\linewidth]{FlowNetS+GRU}{figures/monkaa/18_flownets.png} &
\subfigimg[width=0.32\linewidth]{FlowNetS+Fusion}{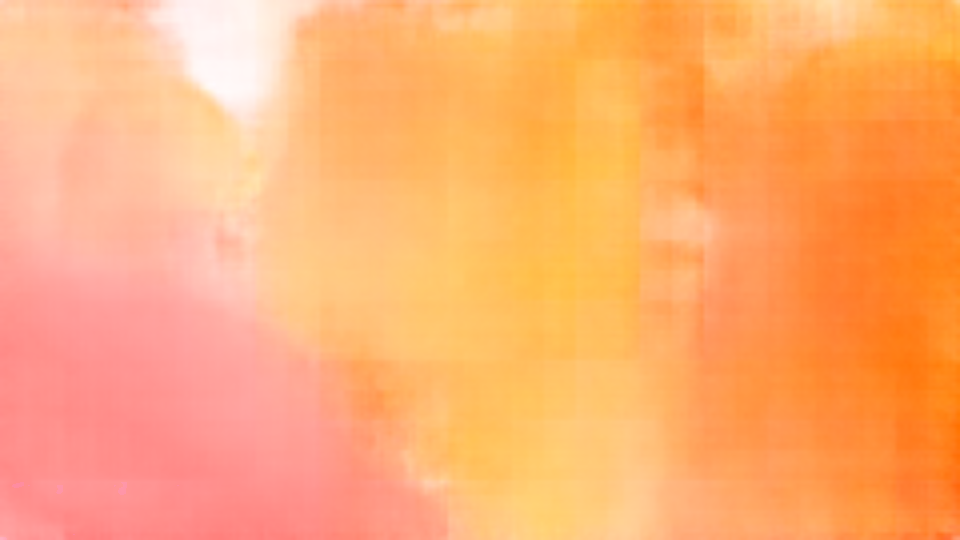} \\
\subfigimg[width=0.32\linewidth]{PWC-Net}{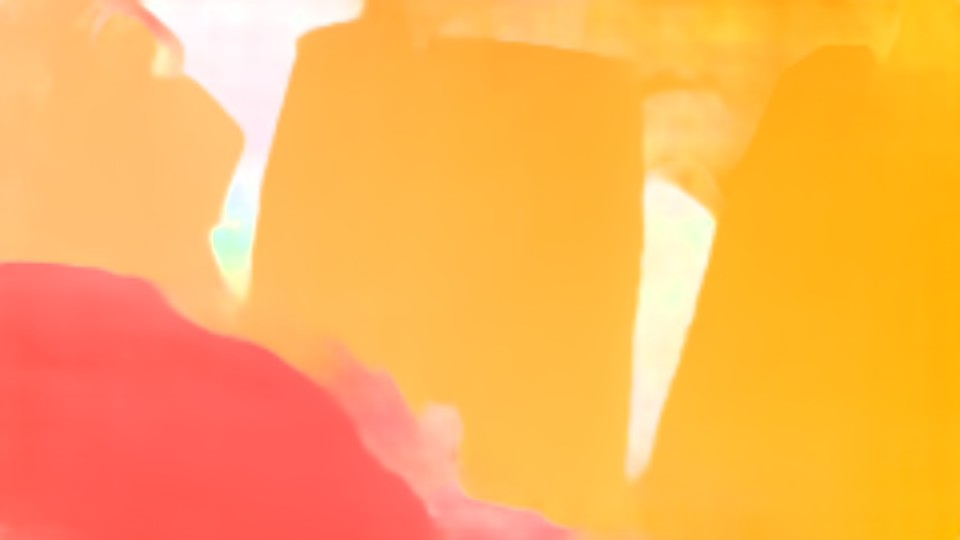} &
\subfigimg[width=0.32\linewidth]{PWC-Net+GRU}{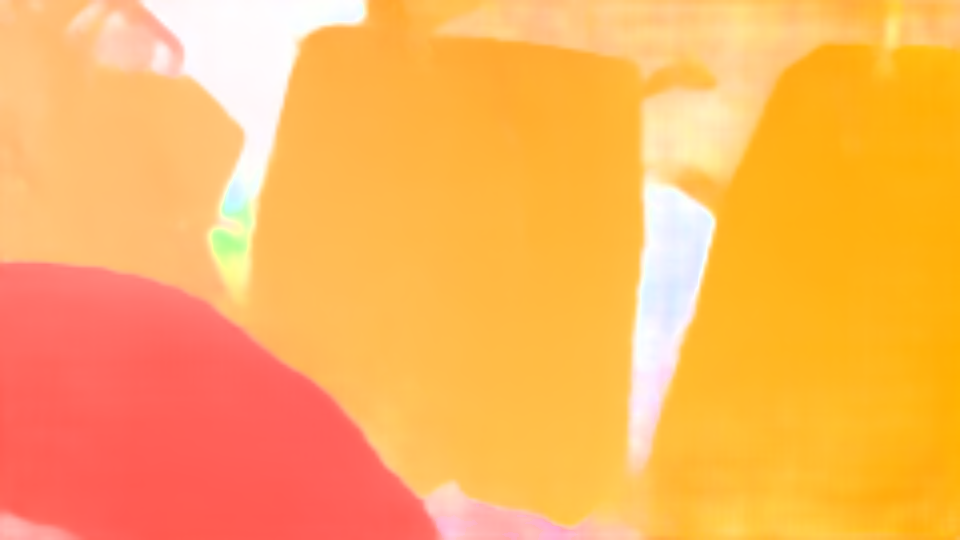} &
\subfigimg[width=0.32\linewidth]{PWC-Net+Fusion}{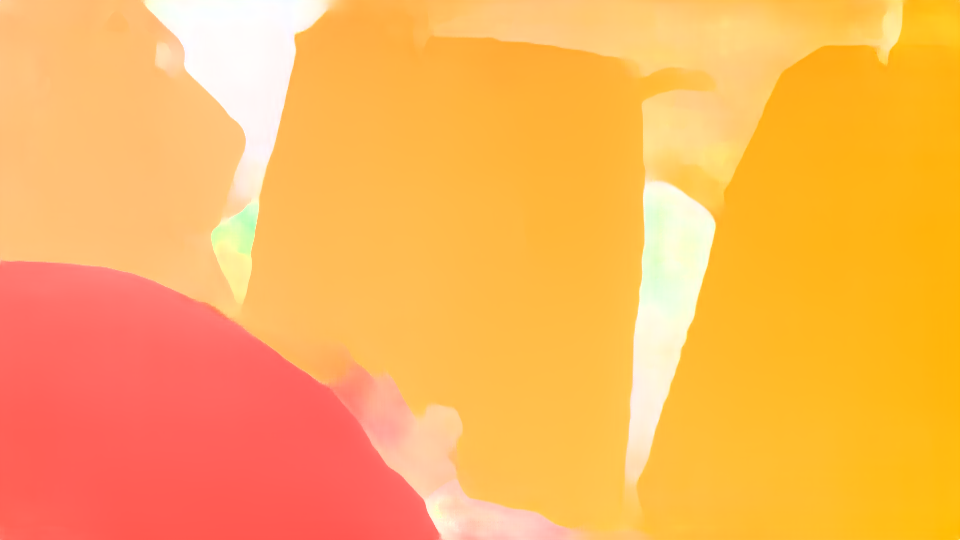} \\

\subfigimg[width=0.32\linewidth]{Input t}{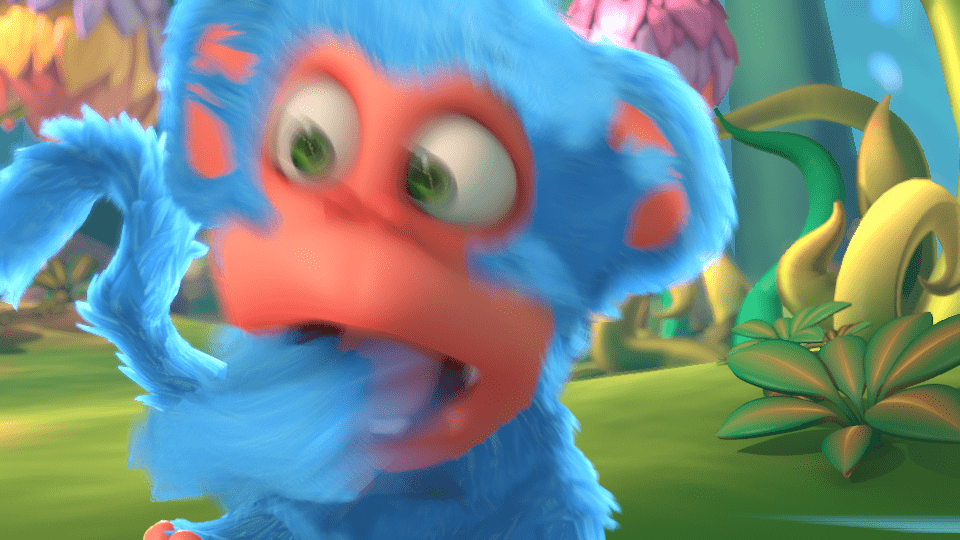} &
\subfigimg[width=0.32\linewidth]{Input t+1}{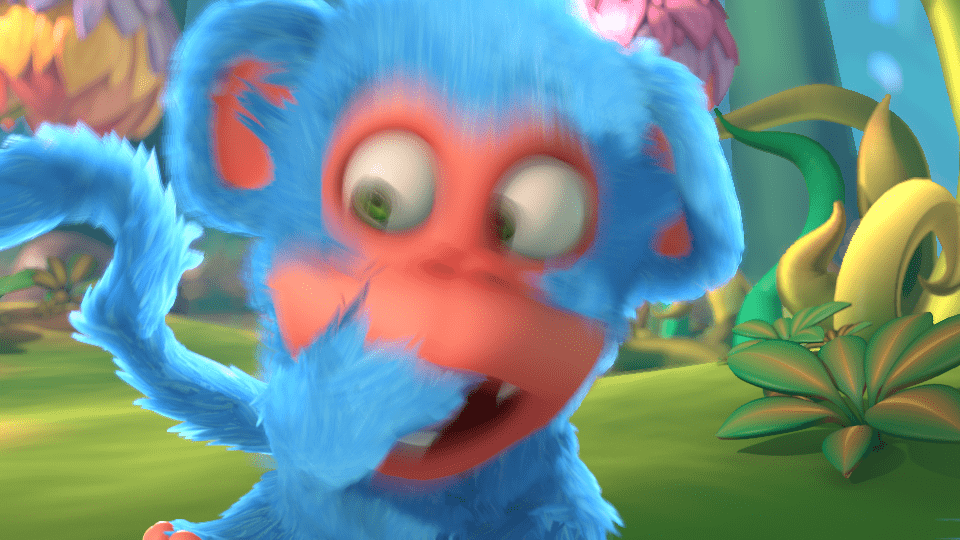} &
\subfigimg[width=0.32\linewidth]{GT Flow}{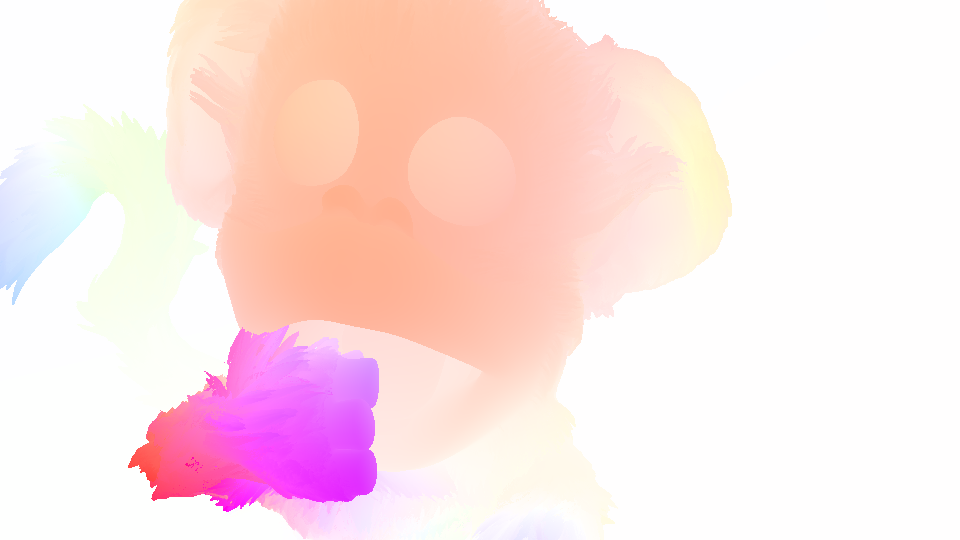} \\
\subfigimg[width=0.32\linewidth]{FlowNetS}{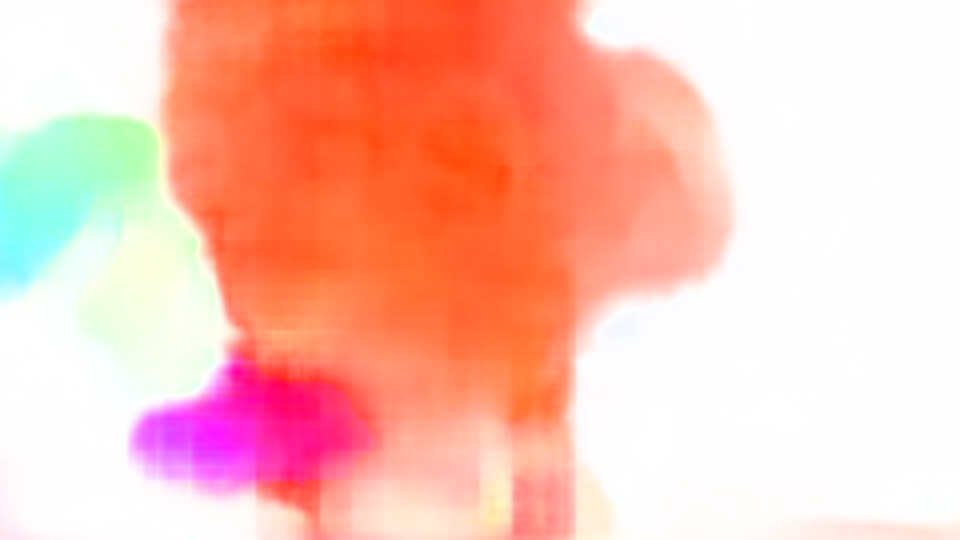} &
\subfigimg[width=0.32\linewidth]{FlowNetS+GRU}{figures/monkaa/40_flownets.png} &
\subfigimg[width=0.32\linewidth]{FlowNetS+Fusion}{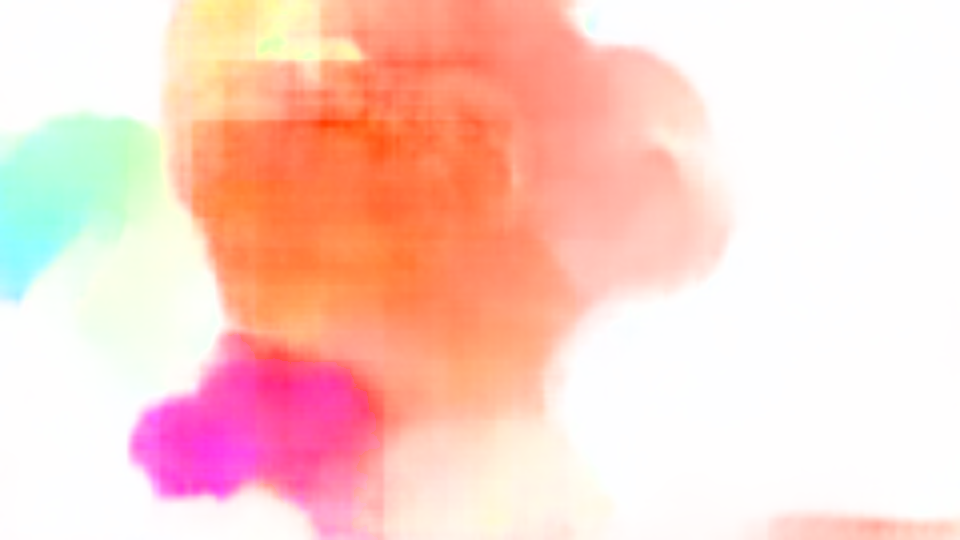} \\
\subfigimg[width=0.32\linewidth]{PWC-Net}{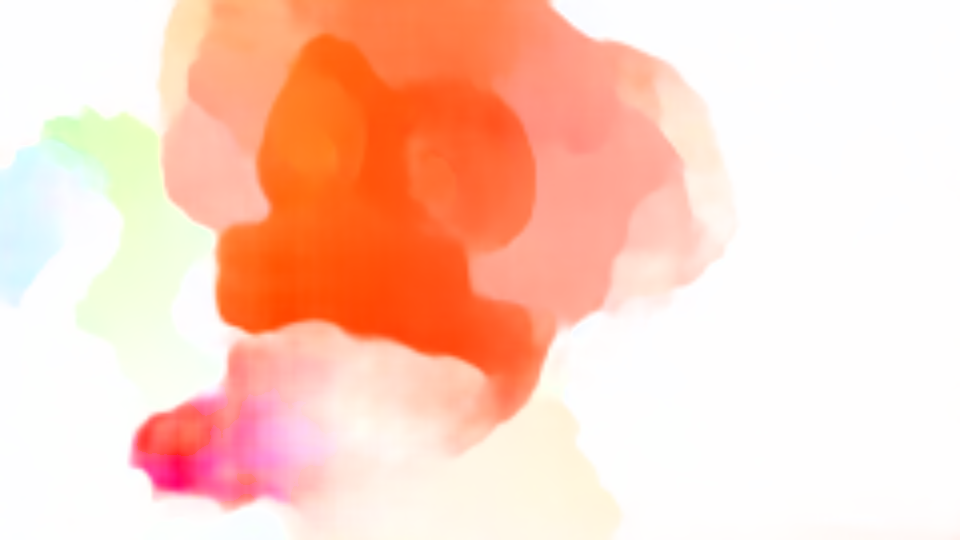} &
\subfigimg[width=0.32\linewidth]{PWC-Net+GRU}{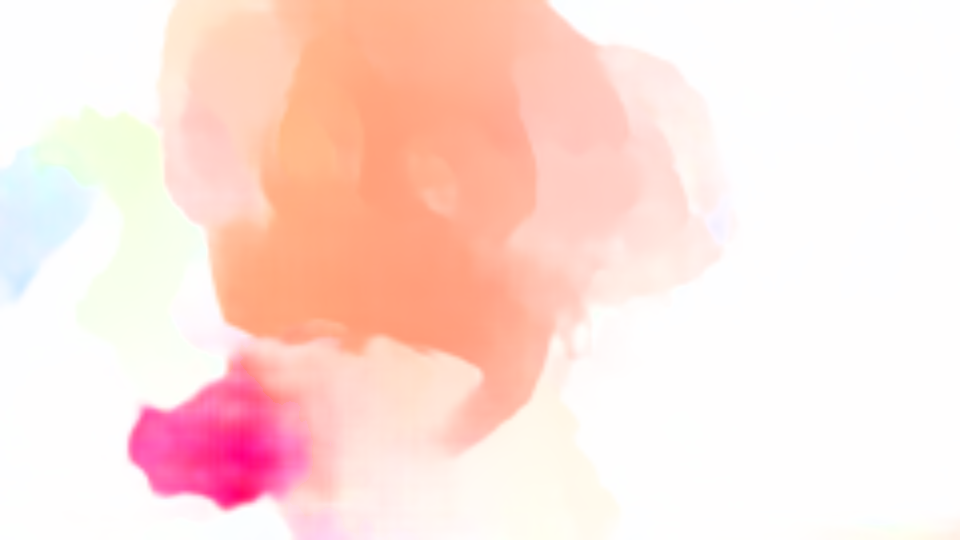} &
\subfigimg[width=0.32\linewidth]{PWC-Net+Fusion}{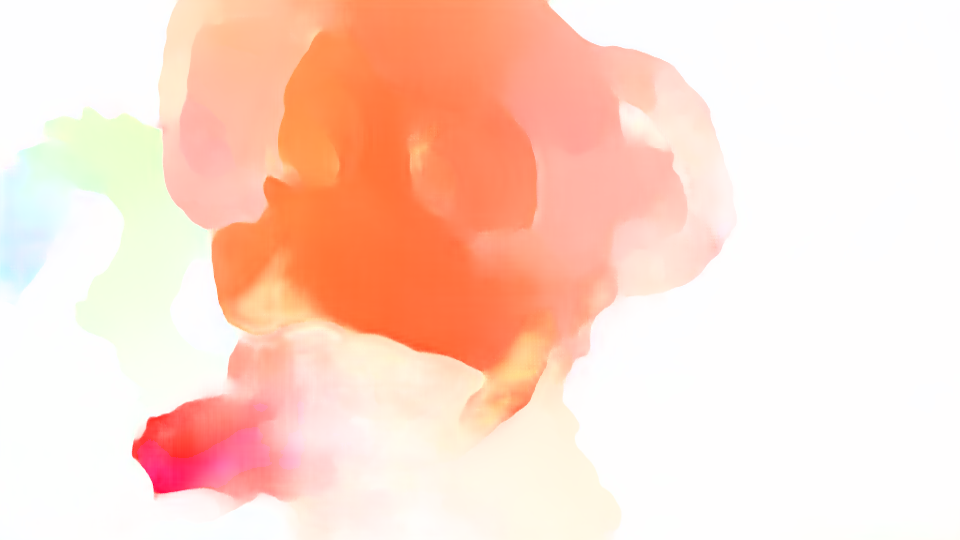} \\
\end{tabular}
\caption{Visualizations of optical flow outputs in the ablation study.}
\vspace{3em}
\label{fig:vis_results_ablataion}
\end{figure*}

The Fusion approach consistently outperforms all other methods, including those using the GRU units. 
This finding suggests that the warping operation can effectively capture the motion trajectory over time, which is missing in the GRU units.
We provide visualizations of the predicted optical flow in the ablation study in Figure~\ref{fig:vis_results_ablataion}.

\subsubsection{The Design of the Fusion Network} 
We also performed an additional ablation study to evaluate the relative importance of each input. 
Our fusion network takes as inputs the optical flow candidates, the brightness errors, and the input image. 
On the validation set of virtual KITTI, its average EPE is \textbf{2.07}. We observe a moderate performance degradation if we  remove the input image (EPE=\textbf{2.15}) or the brightness error (EPE=\textbf{2.16}). 
Adding the flow magnitude to the inputs, on the other hand, leads to almost identical performance (EPE=\textbf{2.09}).
FlowNet2 fuses optical flow predictions of two different networks trained for large and small displacements given the same two frames, and thus flow magnitude is a valuable feature for fusion.
In contrast, we fuse temporal rather than spatial information. The warped and current flows are estimated by the same network, making the magnitude less valuable for fusion. On average, the multi-frame fusion step only takes \textbf{0.02s} in our implementation.

\subsubsection{Using More than Three Frames as Input Helps}
 We also ran oracle experiments using input images with up to 5 frames, and EPE scores are 1.80 (3 frames), 1.64 (4 frames) and 1.56 (5 frames). The results suggest that using more than three frames does  help further, but the relative improvements become smaller.
 In this paper, our main contribution is a deep-learning framework that allows to leverage temporal information. 
Its extension to more than three frames does not require any conceptual modifications, and is a promising area for future research. 

\subsection{The Behavior of the Fusion Network}
We analyzed the average EPE (AEPE) of the flow fields where the fused flow is different than either the current or the warped flow ({\color{red}{red}} regions in Figure~\ref{fig:vis_fusion}).
On the validation set of virtual KITTI, the AEPE of the current flow is 41.88, much higher than 32.71 of the warped flow, suggesting that the latter carries useful information. 
Moreover, the oracle flow has an AEPE of 21.89, confirming that the flow in those regions is incorrectly estimated by both. 
The fused flow has an AEPE of 28.86, lower than both candidate flows but higher than the oracle flow---it does not use the ground truth after all. 
However, we notice that the fused flow has lower errors than the oracle flow for \textbf{37.65\%} of pixels in the red regions, meaning that the estimate is better than any of the inputs.
Again, this is one main advantage of our approach with respect to methods that simply select one of the input flow vectors~\cite{kennedy2015optical}.

\subsection{Comparison with the State-of-the-art}
\begin{figure*}[ht!]
\begin{tabular}{ccc}
\subfigimg[width=0.32\linewidth]{Input t}{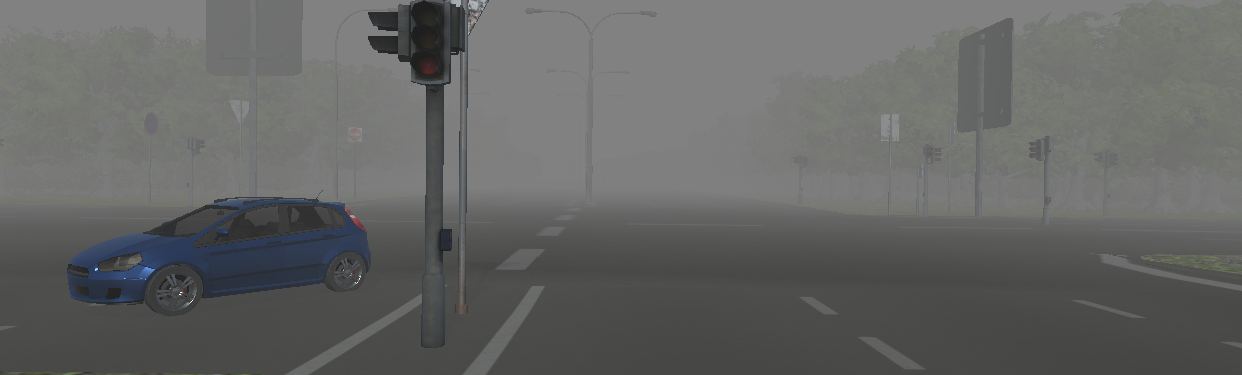} &
\subfigimg[width=0.32\linewidth]{PWC-Net}{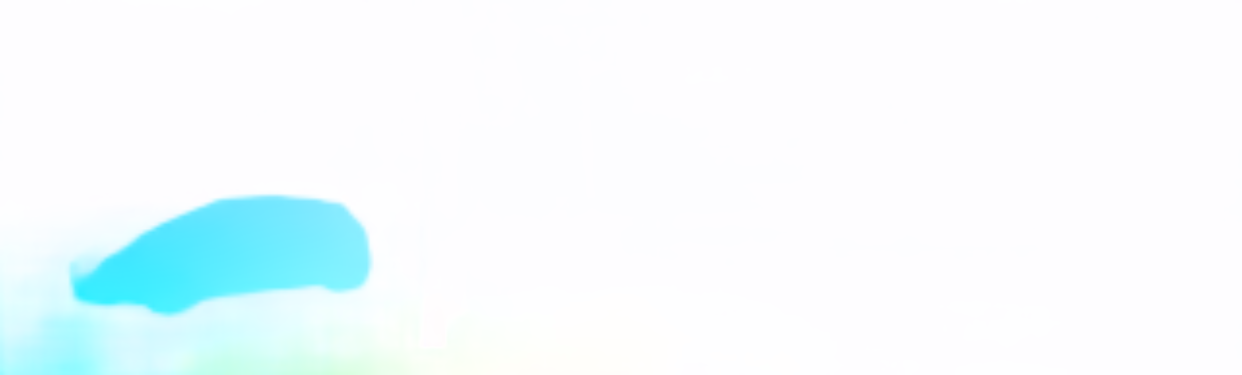} &
\subfigimg[width=0.32\linewidth]{PWC-Net+Fusion}{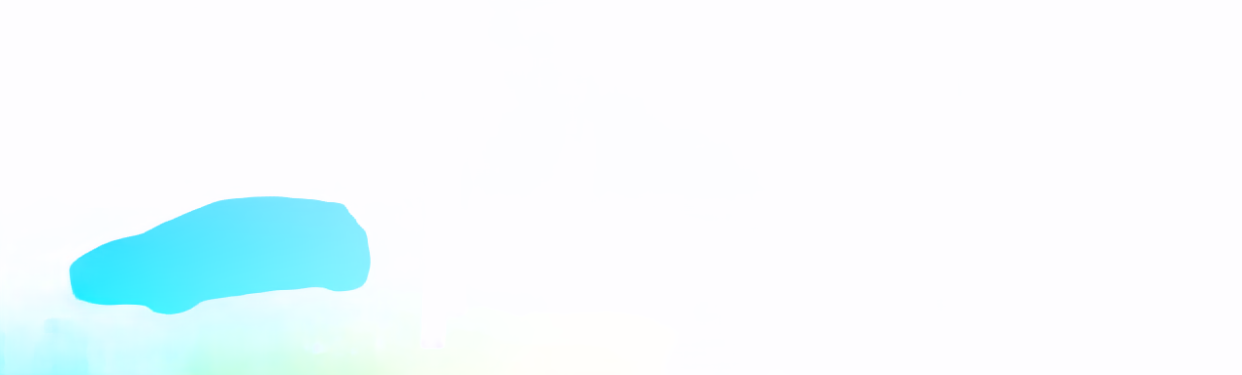} \\
\subfigimg[width=0.32\linewidth]{Error Indication}{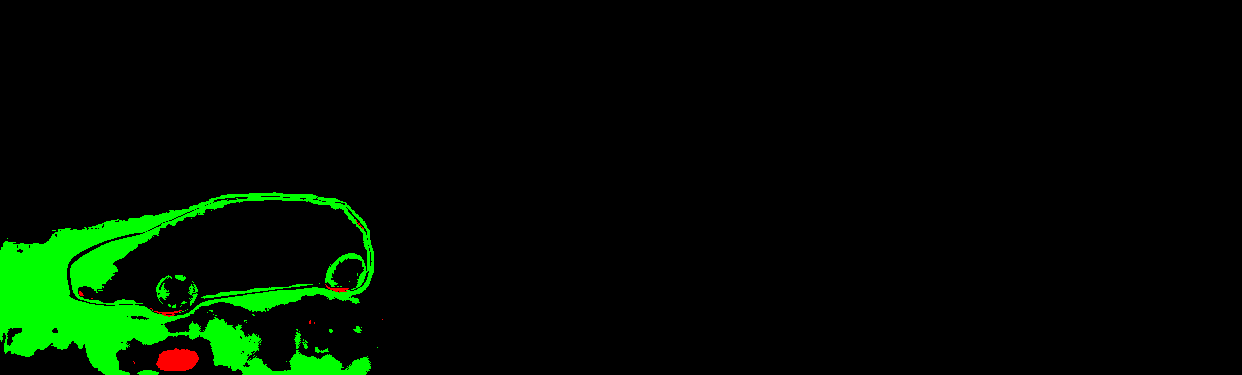} &
\subfigimg[width=0.32\linewidth]{EPE PWC-Net}{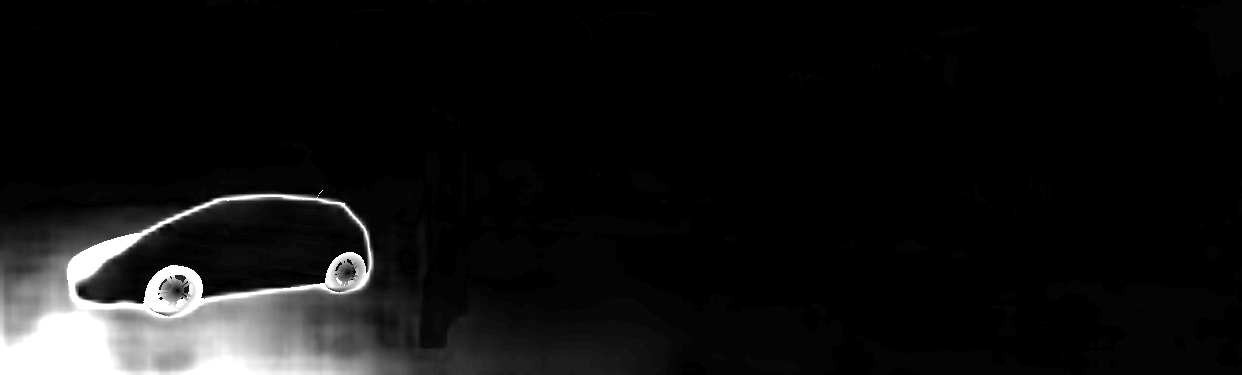} &
\subfigimg[width=0.32\linewidth]{EPE PWC-Net+Fusion}{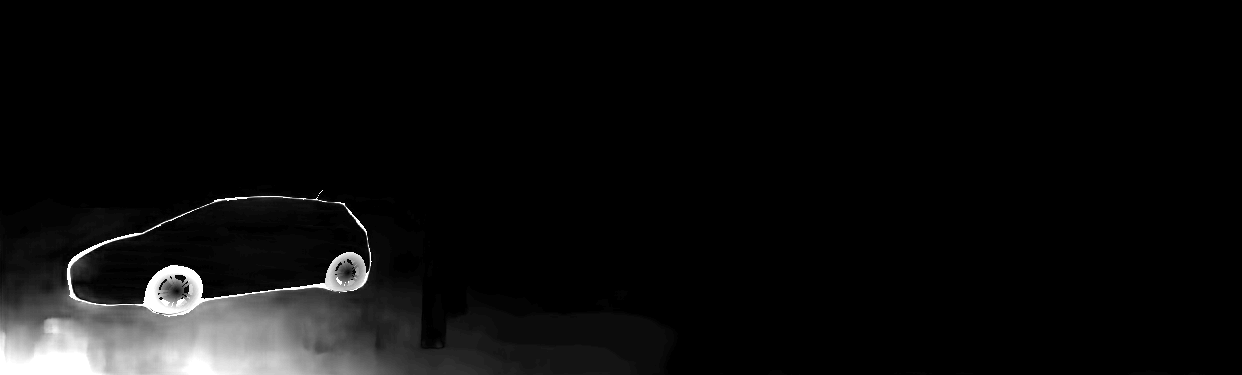} \\
\subfigimg[width=0.32\linewidth]{Input t}{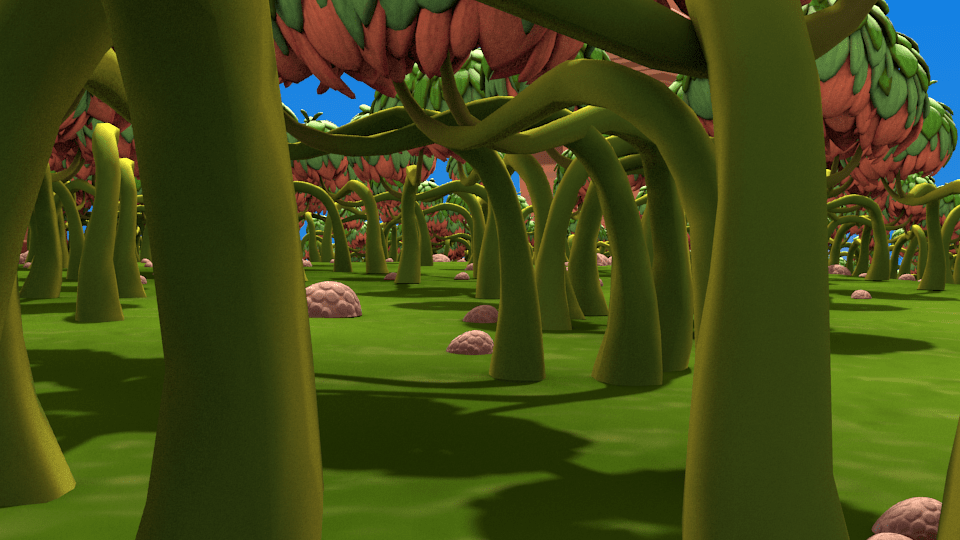} &
\subfigimg[width=0.32\linewidth]{PWC-Net}{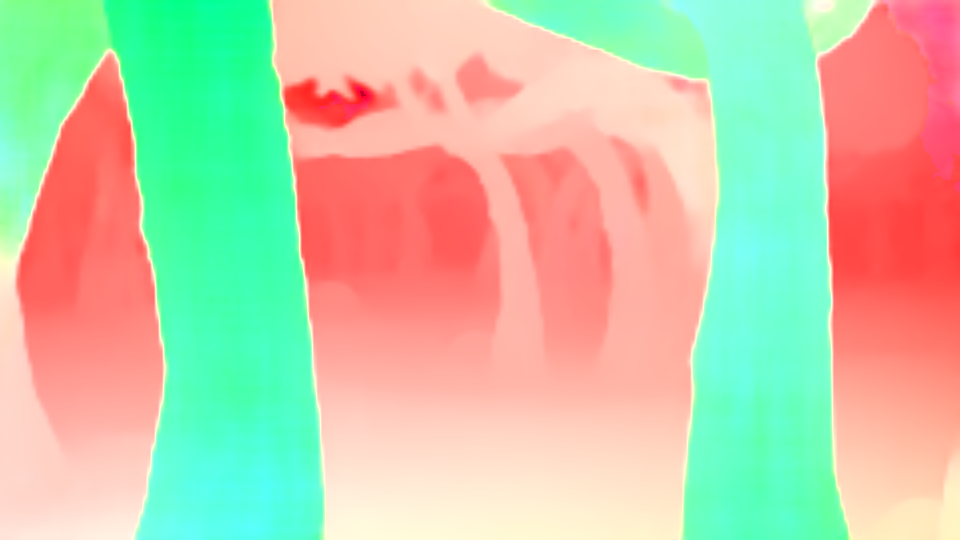} &
\subfigimg[width=0.32\linewidth]{PWC-Net+Fusion}{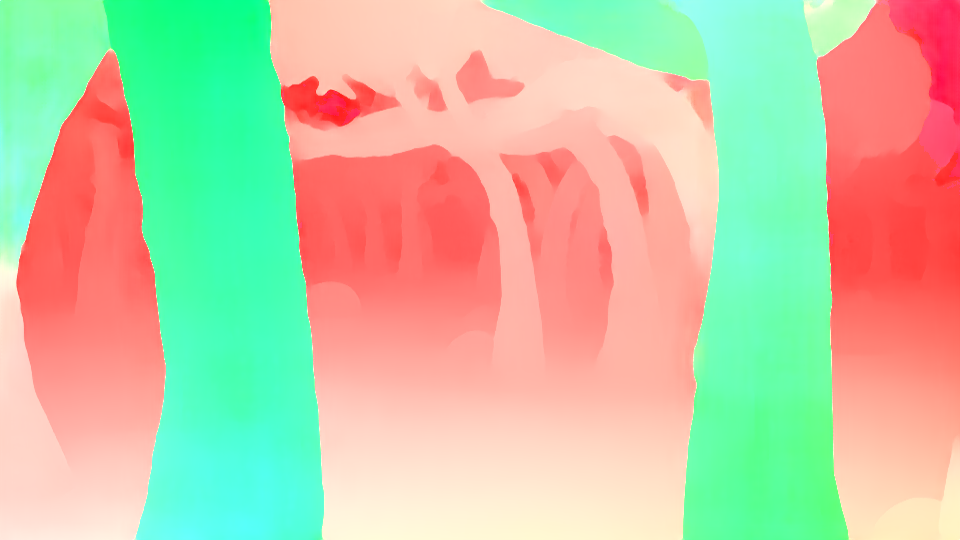} \\
\subfigimg[width=0.32\linewidth]{Error Indication}{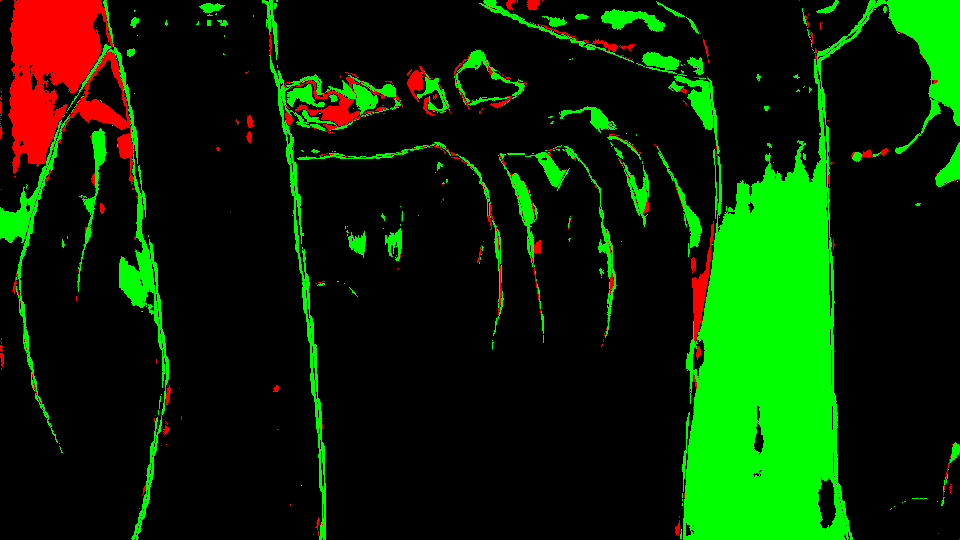} &
\subfigimg[width=0.32\linewidth]{EPE PWC-Net}{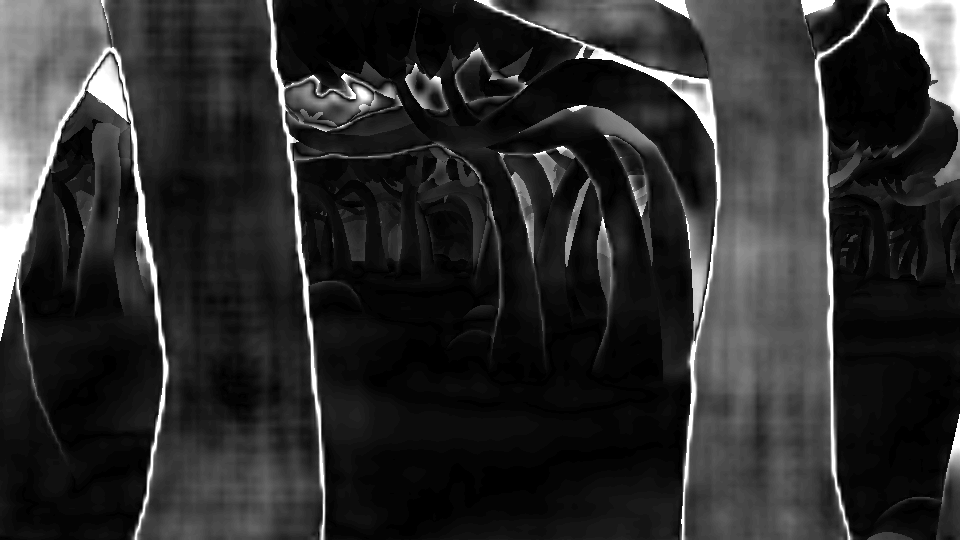} &
\subfigimg[width=0.32\linewidth]{EPE PWC-Net+Fusion}{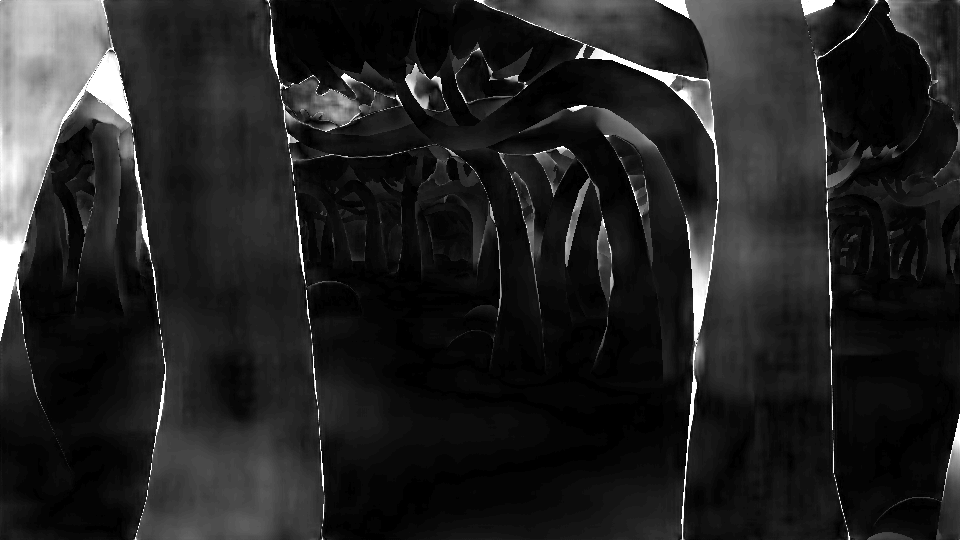} \\
\subfigimg[width=0.32\linewidth]{Input t}{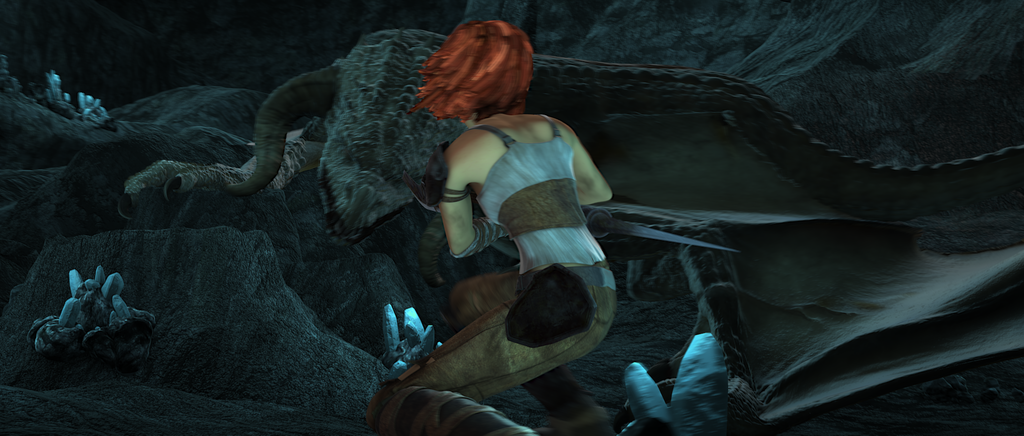} &
\subfigimg[width=0.32\linewidth]{PWC-Net}{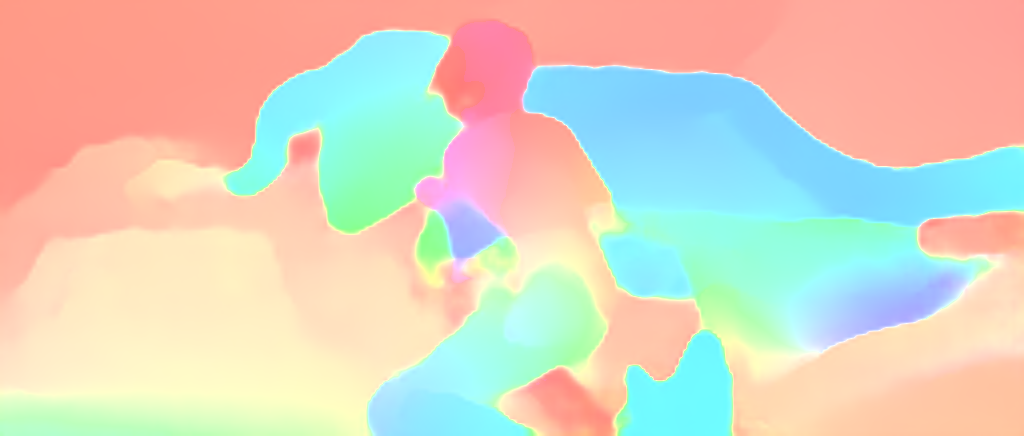} &
\subfigimg[width=0.32\linewidth]{PWC-Net+Fusion}{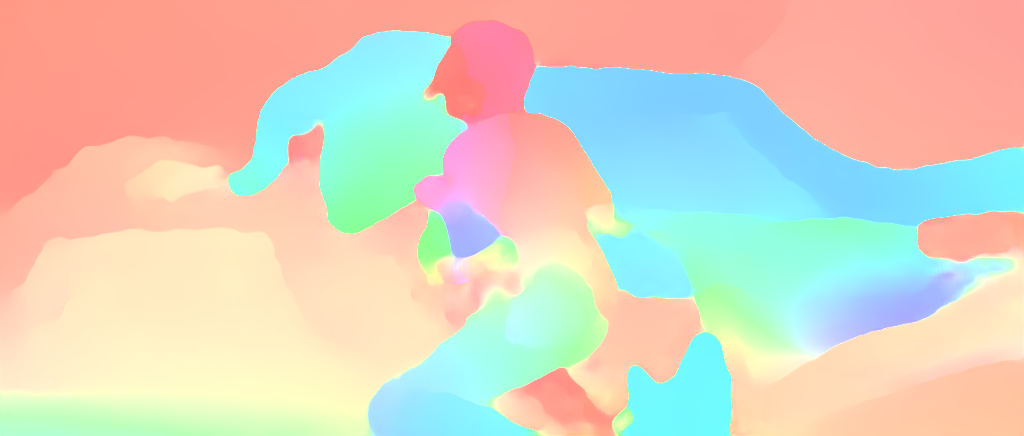} \\
\subfigimg[width=0.32\linewidth]{Error Indication}{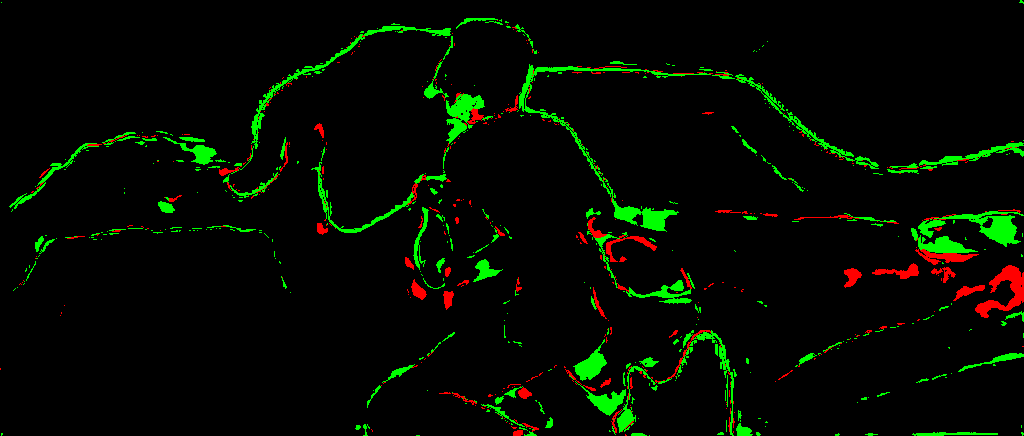} &
\subfigimg[width=0.32\linewidth]{EPE PWC-Net}{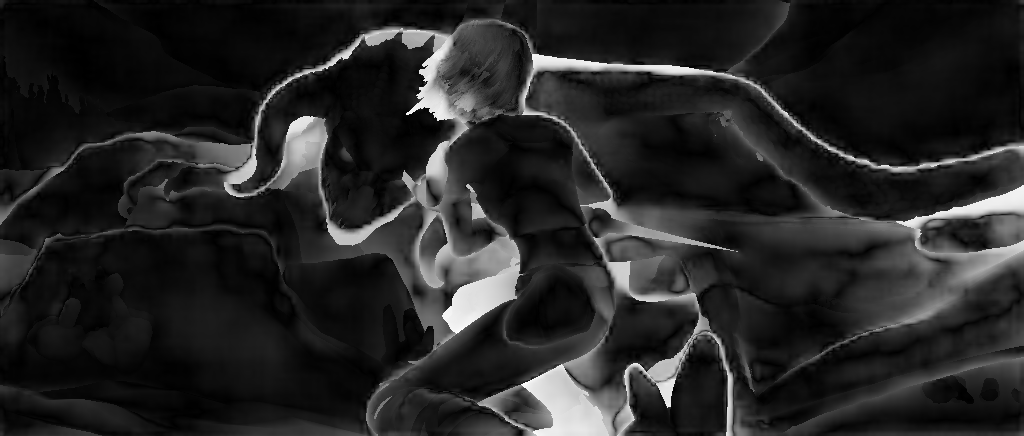} &
\subfigimg[width=0.32\linewidth]{EPE PWC-Net+Fusion}{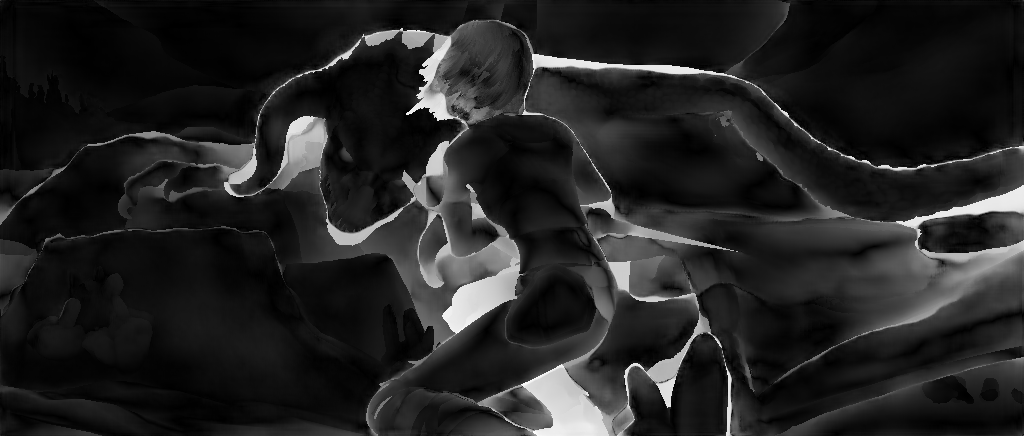} 
\end{tabular}
\caption{Visual results of our fusion method. Our method 
produces consistently better results across different datasets. Notably, our prediction make more
accurate predictions at motion boundaries (green pixels in the indication map mean that PWC-Net+Fusion 
is more accurate than PWC-Net, and red pixels mean that PWC-Net is better).}
\label{fig:vis_results_benchmark}
\end{figure*}

We take the pretrained models of PWC-Net~\cite{sun2017pwc} along with fusion network trained on virtual KITTI 
and Monkaa, and fine-tune the fusion network using the training data from the public benchmarks KITTI~\cite{Geiger:2012:KITTI} and Sintel~\cite{Butler:eccv:2012}.
On the Sintel benchmark (Table~\ref{table:epe_sintel}, left), PWC-Fusion consistently outperforms PWC-Net~\cite{sun2017pwc}, particularly in unmatched (occlusion) regions. Its performance is also better
than a recent multi-frame optical flow estimation method~\cite{maurer2018proflow}.
On the KITTI benchmark (Table~\ref{table:epe_sintel}, right), PWC-Fusion outperforms all two-frame optical flow methods
including the state-of-the-art PWC-Net. This is also the first time a multi-frame optical flow algorithm consistently
outperforms two-frame approaches across different datasets. We provide some visual results in Figure~\ref{fig:vis_results_benchmark},
and provide indicator maps to demonstrate the behavior of fusion network.

\section{Conclusions}\label{sec:conclusions}
We have presented a simple and effective fusion approach for multi-frame optical flow estimation. Our main observation is that multiple frames provide new information beyond what is available only looking at two adjacent frames, in particular for occluded and out-of-boundary pixels. Thus we proposed to fuse the warped previous optical flow with the current optical flow estimate. Extensive experiments demonstrate the benefit of our fusion approach: it outperforms both two-frame baselines and sensible multi-frame baselines based on GRUs. Moreover, it is top-ranked among all published flow methods on the MPI Sintel and KITTI 2015 benchmark. 

While the performance of two-frame methods has been steadily pushed over the years,  we begin to see diminishing returns. Our work shows that it could be fruitful to exploit temporal information for optical flow estimation and we hope to see more work in this direction.

\paragraph{Acknowledgments}
We thank Fitsum Reda and Jinwei Gu for their help with the software implementation, Xiaodong Yang for the helpful discussions about RNN models, and Simon Baker for providing insights about flow fusion.

{\small
\bibliographystyle{ieee}
\bibliography{main}
}

\end{document}